\documentclass{article}

\usepackage[preprint,nonatbib]{neurips_2021}

\usepackage[utf8]{inputenc} 
\usepackage[T1]{fontenc}    
\usepackage{hyperref}       
\usepackage{url}            
\usepackage{booktabs}       
\usepackage{amsfonts}       
\usepackage{nicefrac}       
\usepackage{microtype}      

\usepackage{amsmath}
\usepackage{graphicx}
\usepackage{algorithm}
\usepackage{algorithmic}
\usepackage{multirow}
\usepackage{subcaption}
\usepackage{todonotes}
\usepackage{xfrac}

\newcommand{\eg}{\textit{e.g.}}
\newcommand{\ie}{\textit{i.e.}}
\newcommand{\aka}{\textit{a.k.a.}}
\newcommand{\etal}{\textit{et al.}}
\newcommand{\ours}{EBM-Fold}

\title{\ours{}: Fully-Differentiable Protein Folding Powered by Energy-based Models}

\author{
  Jiaxiang Wu \\
  Tencent AI Lab \\
  \texttt{jonathanwu@tencent.com} \\
  \And
  Shitong Luo \\
  Peking University \\
  \texttt{luost@pku.edu.cn} \\
  \And
  Tao Shen \\
  Tencent AI Lab \\
  \texttt{scotttshen@tencent.com} \\
  \And
  Haidong Lan \\
  Tencent AI Lab \\
  \texttt{haidonglan@tencent.com} \\
  \And
  Sheng Wang \\
  Tencent AI Lab \\
  \texttt{shengwwang@tencent.com} \\
  \And
  Junzhou Huang \\
  Tencent AI Lab \\
  \texttt{joehhuang@tencent.com} \\
}

\begin{document}

\maketitle

\begin{abstract}

Accurate protein structure prediction from amino-acid sequences is critical to better understanding proteins' function. Recent advances in this area largely benefit from more precise inter-residue distance and orientation predictions, powered by deep neural networks. However, the structure optimization procedure is still dominated by traditional tools, \eg{} Rosetta, where the structure is solved via minimizing a pre-defined statistical energy function (with optional prediction-based restraints). Such energy function may not be optimal in formulating the whole conformation space of proteins. In this paper, we propose a fully-differentiable approach for protein structure optimization, guided by a data-driven generative network. This network is trained in a denoising manner, attempting to predict the correction signal from corrupted distance matrices between $C_{\alpha}$ atoms. Once the network is well trained, Langevin dynamics based sampling is adopted to gradually optimize structures from random initialization. Extensive experiments demonstrate that our \ours{} approach can efficiently produce high-quality decoys, compared against traditional Rosetta-based structure optimization routines.
\end{abstract}

\section{Introduction}

The biological function of a protein is largely determined by its 3-dimensional structure \cite{roy2012cofactor}. Protein structure determination through experimental methods, \eg{} NMR, crystallography X-Ray, and cryo-EM, is able to produce highly accurate 3D structures, but can be quite time consuming in practice and may not be applicable for all the proteins (\eg{} membrane proteins). On the other hand, computational approaches for protein structure prediction from amino-acid sequences have been actively explored for the past few decades, but most methods' prediction accuracy is still far from satisfactory. Nevertheless, this remains an important problem due to its critical role in various applications, including protein design and structure-based drug discovery.

In 2020, the CASP14 competition\footnotemark{} (\textbf{C}ritical \textbf{A}ssessment of Techniques for Protein \textbf{S}tructure \textbf{P}rediction) witnessed a major breakthrough in this area. DeepMind proposed the AlphaFold2 algorithm\cite{jumper2020high}, which produced highly accurate structure predictions (GDT-TS over 0.9) for over two-thirds of CASP14 target proteins. Under such a GDT-TS threshold, AlphaFold2 achieved the atomic-level resolution for protein structure prediction, which is comparable or sometimes even better than the experimental data. AlphaFold2 built an attention-based neural network system, which allowed end-to-end training from multiple sequence alignments (MSAs) to protein structures. More technical details of AlphaFold are not yet published at the moment (May 2021), but there are already lots of interests and attempts in understanding and reimplementing it \cite{alquraishi2020alphafold2, service2020game, fuchs2021iterative, wang2021alphafold2}.

\footnotetext{\url{https://predictioncenter.org/casp14}}

In this paper, we propose a fully-differentiable structure optimization module to demonstrate the possibility of replacing traditional protein structure optimization tools, \eg{} Rosetta \cite{simons1999ab} and I-TASSER \cite{roy2010tasser}. Previously, protein structure predictions heavily rely on these tools to build 3D structures under the guidance of a statistic energy function and/or structural templates, optionally assisted by inter-residue distance and/or orientation predictions \cite{xu2019distance, yang2020improved}. It is non-trivial to build a end-to-end structure prediction pipeline based on these tools, as their structure optimization process often involves multiple stages and is not fully-differentiable. Here, we present an alternative choice for building 3D structures from inter-residue distance and orientation predictions. Our approach, namely \ours{}, employs an energy-based model to implicitly learn the underlying data distribution of native protein structures. The \ours{} approach is fully-differentiable, and is able to gradually refine randomly initialized structure predictions to high accuracy. This could motivate future attempts in building an end-to-end system for protein structure prediction, similar as AlphaFold2.

Specifically, \ours{} trains a neural network to estimate native protein structures' data distribution's log probability's gradients over atom coordinates. This network is trained in a denoising manner, \ie we randomly perturb native structures' atom coordinates as the network's inputs, and force the network to estimate the corresponding gradients over perturbed inputs. During the inference phase, we randomly initialize protein structures without any restraints, and gradually refine these structures using the network's estimated gradients over atom coordinates. For now, we only consider the prediction of each residue's $C_{\alpha}$ atom's 3D coordinate, but it is straightforward to extend the current method to predict all the atoms' 3D coordinates in a protein.

Our \ours{} approach is light-weighted and computational efficient. It takes around 18 hours to train the network on a single Nvidia V100 GPU card, and takes less than 1 minute to produce one predicted structure for proteins shorter than 300 residues. We validate the \ours{}'s effectiveness on a medium-scale domain-level dataset (about 2000 domains in total for train/valid/test), following CATH domain definitions \cite{sillitoe2020cath}. We use trRosetta \cite{yang2020improved} as the baseline method, which is one of the state-of-art-methods for protein structure optimization. Under the same inter-residue distance and orientation predictions, \ours{} improves the averaged lDDT-Ca score \cite{mariani2013lddt} from 0.7452 (trRosetta) to 0.7775, computed on all the predicted structures of each method. However, when considering the optimal structure prediction for each target, \ours{} is inferior to trRosetta (0.8124 vs. 0.8523), indicating that \ours{} still needs further improvements.

There has been a few early works in adopting energy-based models for molecular conformation generation \cite{shi2021learning} and protein structure prediction \cite{ingraham2019learning, du2020energy}. Both \cite{ingraham2019learning, du2020energy} explicitly learn an energy function to measure the probability distribution's log likelihood, and use its gradients over input features/coordinates as update signals for structure optimization. In contrast, we implicitly models the data distribution with a score network, which directly outputs estimated gradients over 3D coordinates for protein structure prediction. On the other hand, Shi et al. \cite{shi2021learning} propose to learn a score network via denoising score matching \cite{song2019generative} to estimate gradients over the inter-atom distance for small molecules, and then propagate these gradients to per-atom 3D coordinates via the chain rule. Such gradients are equivariant to 3D translation and rotation, and thus can be used to gradually refine molecular structures from random initialization. Their method is highly effective in predicting 3D conformations for small molecules (tens of atoms), but its extension for macro-molecules (\eg{} proteins), which often consists of over hundreds of amino-acids and thousands of atoms, is not yet explored.

\section{Related Work}

\textbf{Protein structure prediction.} Early attempts in predicting a protein's structure from its amino-acid sequence mostly rely on statistics energy functions \cite{omeara2015combined, park2016simultaneous}, structural templates \cite{song2013high}, and fragment assembly techniques \cite{simons1997assembly}. To introduce additional restraints to assist the structure optimization process, co-evolutional information (\eg{} multiple sequence alignment) can be exploited to predict distance and orientation between different residues. Binary-valued contact predictions ($C_{\beta}$-$C_{\beta}$ atom pairs closer than 8 Angstrom) are firstly used in \cite{adhikari2015confold, adhikari2018confold2, hou2019protein, gao2019destini}. With the growing prediction ability of deep neural networks, real-valued distance of atom pairs can also be predicted and converted into differentiable energy terms, used along with traditional statistics energy functions \cite{xu2019analysis, xu2019distance, hou2020multicom}. In \cite{senior2019protein, senior2020improved, yang2020improved}, inter-residue orientation (dihedral and plane angles) are further cooperated to non-redundant restraints besides ones derived from distance predictions. Still, most of these methods depends on a pre-defined statistic energy function, and restraints derived from neural network's predictions only serve as additional restraints/energy terms.

There are a few works in exploring alternative approaches for protein structure optimization. Alipanahi et al. \cite{alipanahi2013determining} propose an semedefinite programming (SDP) approach to determine protein structures from noisy distance constraints. Anand et al. \cite{anand2018generative} adopt an ADMM solver to recover structures from the generated pairwise distance matrix for protein design.

AlphaFold2 \cite{jumper2020high} is the first end-to-end approach, to the best of our knowledge, for building protein structures directly from the co-evolution information, represented as multiple sequence alignments. It seems that they adopt a SE(3)-equivariant model, \eg{} SE(3)-Transformer \cite{fuchs2020se3} and LieTransformer \cite{hutchinson2021lietransformer}, to iteratively optimize the protein structure, under the guidance of learnt embeddings of inter-residue relationships. Still, much of AlphaFold2's technical details remain unclear at the moment.

\textbf{Energy-based models.} Energy-based models (EBMs) provide a less restrictive way in formulating a unknown data distribution. These models define an energy function as the unnormalized negative log probability, which can be parameterized by arbitrary regression function, \eg{} neural networks. To sample new data from this distribution, one can employ the Langevin dynamics sampling process \cite{teh2003energy, xie2016theory}. Score matching \cite{hyvarinen2005estimation} provides an alternative way to approximate the data distribution, without the time-consuming MCMC sampling in the training process. For a more comprehensive review of energy-based models, please refer to \cite{song2021train}.

Energy-based models have been applied in various domains, including image generation \cite{xie2016theory, du2019implicit, song2019generative, song2020improved}, video generation \cite{xie2017synthesizing, xie2021learning}, 3D shape pattern generation \cite{xie2018learning, xie2020generative}, 3D point cloud generation \cite{xie2021generative}, density estimation \cite{wenliang2019learning, song2020sliced}, and reinforcement learning \cite{haarnoja2017reinforcement}. Nevertheless, the application of energy-based models in protein structure prediction is much less explored. Du \etal{} \cite{du2020energy} propose to learn a data-driven energy function via EBM, and demonstrates its effectiveness in side-chain conformation prediction. NEMO \cite{ingraham2019learning} performs iterative refinement over internal coordinates based protein representations via a unrolled Langevin dynamics simulator. However, the co-evolution information is not exploited, which is critical to producing accurate protein structure predictions.

\section{Preliminaries}

In this section, we firstly introduce basic concepts of energy-based models and their connection with protein structure optimization. Afterwards, we describe how to adopt deep neural networks to predict inter-residue distance and orientations, which will serve as one of the critical input features for the upcoming structure optimization process.

\subsection{Energy-based Models}

Energy-based models (EBMs) aim at obtaining an energy function, mostly parameterized by neural networks, to assign low energies to inputs drawn from the true data distribution, and high energies to the others. Such energy function can be used to sample new data following its corresponding probability distribution via Langevin dynamics \cite{du2019implicit}.

Formally, the energy function is defined as $E_{\theta} \left( \mathbf{x} \right) \in \mathbb{R}$ where $\mathbf{x} \in \mathbb{R}^{d}$ is the input and $\theta$ represents the energy function's parameters (\eg{} a neural network's weights). The corresponding probability distribution can be written as:
\begin{equation}
p_{\theta} \left( \mathbf{x} \right) = \frac{\exp \left[ - E_{\theta} \left( \mathbf{x} \right) \right]}{\int \exp \left[ - E_{\theta} \left( \mathbf{x} \right) \right] d \mathbf{x}}
\label{eqn:ebm_prob_def}
\end{equation}
where the denominator (\aka{} the partition function) is a integral over the whole input space, which is independent of the input $\mathbf{x}$. In order to sample new data from the above probability distribution, one may employ Langevin dynamics to iteratively update the data from random initialization:
\begin{equation}
\mathbf{x}_{t} := \mathbf{x}_{t - 1} - \frac{\lambda_{t}}{2} \cdot \nabla_{\mathbf{x}} E_{\theta} \left( \mathbf{x}_{t - 1} \right) + \sqrt{\lambda_{t}} \cdot \mathbf{v}_{t}
\label{eqn:ebm_ld_sampling}
\end{equation}
where $\lambda_{t}$ is the step-size at the $t$-th iteration, and $\mathbf{v}_{t} \in \mathcal{N} \left( \mathbf{0}, \mathbf{I} \right)$. Intrinsically, $\mathbf{x}_{t}$ is updated with the energy function's gradients, together with an additive random noise. In \cite{welling2011bayesian}, authors showed that if $t \rightarrow \infty$ and $\lambda_{t} \rightarrow 0$, then the above iterative update process can generate data samples following the probability distribution $p_{\theta} \left( \mathbf{x} \right)$.

Song et al. \cite{song2019generative} proposed a more direct approach to enable Langevin dynamics sampling from EBMs. Since the partition function in Eq. (\ref{eqn:ebm_prob_def}) is independent to $\mathbf{x}$, we have $\nabla_{\mathbf{x}} \log p_{\theta} \left( \mathbf{x} \right) = - \nabla_{\mathbf{x}} E_{\theta} \left( \mathbf{x} \right)$. Hence, the Langevin dynamics sampling process can be rewritten as:
\begin{equation}
\begin{split}
\mathbf{x}_{t} :=&~ \mathbf{x}_{t - 1} + \frac{\lambda_{t}}{2} \cdot \nabla_{\mathbf{x}} \log p_{\theta} \left( \mathbf{x}_{t - 1} \right) + \sqrt{\lambda_{t}} \cdot \mathbf{v}_{t} \\
=&~ \mathbf{x}_{t - 1} + \frac{\lambda_{t}}{2} \cdot h_{\theta} \left( \mathbf{x}_{t - 1} \right) + \sqrt{\lambda_{t}} \cdot \mathbf{v}_{t}
\end{split}
\label{eqn:ebm_ld_sampling_v2}
\end{equation}

Therefore, instead of explicitly learning an energy function $E_{\theta} \left( \mathbf{x} \right)$, they propose to build a score network $h_{\theta} \left( \mathbf{x} \right) \in \mathbb{R}^{d}$ to directly estimate gradients of the true data distribution's log probability, given by $\nabla_{\mathbf{x}} \log p_{d} \left( \mathbf{x} \right)$. The score network is trained by randomly perturbing the original data with various levels of random noises, and then use corresponding gradients as the supervised information. The optimization objective is given by:
\begin{equation}
\frac{1}{2 K} \sum_{k = 1}^{K} \sigma_{k}^{2} \cdot \mathbb{E}_{p_{d} \left( \mathbf{x} \right) q \left( \tilde{\mathbf{x}} | \mathbf{x}, \sigma_{k} \right)} \left\| h_{\theta} \left( \mathbf{x} \right) - \frac{\mathbf{x} - \tilde{\mathbf{x}}}{\sigma_{k}^{2}} \right\|_{2}^{2}
\label{eqn:ebm_dsm_loss_func}
\end{equation}
where the perturbed data $\tilde{\mathbf{x}}$ is generated by adding a random noise drawn from $\mathcal{N} \left( \mathbf{0}, \sigma_{k}^{2} \mathbf{I} \right)$ to the original data $\mathbf{x}$. Such perturbation scheme leads to a quite concise form of log probability's gradients (the second term within the above $l_{2}$-norm operator). It is worth noting that the score network often conditions on the random noise's standard deviations $\left\{ \sigma_{k} \right\}$, since this information is critical in estimating the gradient signal. Once the score network is trained, it can be used to perform Langevin dynamics sampling to generate new data samples, as in Eq. (\ref{eqn:ebm_ld_sampling_v2}).

Here, we note that there is indeed a close connection between energy-based models and protein structure optimization. Given an amino-acid sequence $s$, the protein structure optimization problem can be formulated as finding the most-likely 3D structure corresponding to this sequence. Let $p_{d} \left( \mathbf{x} | s \right)$ denote the conditional probability distribution of all the possible 3D structures, then this problem can be solved in a two-stage manner: 1) train a EBM to approximate the true data distribution $p_{d} \left( \mathbf{x} | z \right)$; and 2) run Langevin dynamics to gradually update a randomly initialized structure to fit the approximated data distribution $p_{\theta} \left( \mathbf{x} | z \right)$. This forms the core motivation of our \ours{} approach.

\subsection{Distance \& Orientation Prediction}

Traditionally, protein structures are optimized with respect to a pre-defined statistics energy function, \eg{} Rosetta's ``talaris2014'' and ``ref2015'' score functions \cite{omeara2015combined, park2016simultaneous}. However, such energy functions alone may not be sufficient to produce high-quality structures due to insufficient exploration in a huge conformation space.

To tackle this dilemma, many works \cite{adhikari2015confold, xu2019distance, senior2019protein, senior2020improved} firstly predict the distance and/or orientation (dihedral and plane angles) between different residues in the protein, and then convert them into additional energy terms to measure the consistency between structures under optimization and distance \& orientation predictions. These energy terms are combined with the traditional statistics energy function and jointly used for structure optimization.

Among these works, Yang et al. \cite{yang2020improved} proposed the trRosetta method, consists of a deep residual network to jointly predict inter-residue distance and orientation and a Rosetta-based structure optimization protocol to fully exploit such predictions, which achieved the state-of-the-art accuracy for structure optimization. In trRosetta, four types of inter-residue relationships are considered:
\begin{itemize}
\item $d_{ij}$: distance defined by $C_{\beta}^{( i )}$-$C_{\beta}^{( j )}$ atoms
\item $\omega_{ij}$: dihedral angle defined by $C_{\alpha}^{( i )}$-$C_{\beta}^{( i )}$-$C_{\beta}^{( j )}$-$C_{\alpha}^{( j )}$ atoms
\item $\gamma_{ij}$: dihedral angle defined by $N^{( i )}$-$C_{\alpha}^{( i )}$-$C_{\beta}^{( i )}$-$C_{\beta}^{( j )}$ atoms
\item $\varphi_{ij}$: plane angle defined by $C_{\alpha}^{( i )}$-$C_{\beta}^{( i )}$-$C_{\beta}^{( j )}$ atoms
\end{itemize}
where $N^{( i )}$, $C_{\alpha}^{( i )}$, and $C_{\beta}^{( i )}$ are the $i$-th residue's $N$, $C_{\alpha}$, and $C_{\beta}$ atoms, respectively.

The inter-residue distance and orientation predictor uses various features extracted from the multiple sequence alignment (MSA), and pack them into a $L \times L \times D_{in}$ tensor as input feature maps, where $L$ is the sequence length and $D_{in}$ is the number of input feature dimensions. The prediction task is formulated as a pixel-wise classification problem, where distance and angle values are discretized into bins and the network produces probabilistic predictions. Here, we denote the number of bins for the above four types of inter-residue relationships as $D_{d}$, $D_{\omega}$, $D_{\gamma}$, and $D_{\varphi}$. Thus, the inter-residue distance and orientation predictor's final outputs of size $L \times L \times \left( D_{d} + D_{\omega} + D_{\gamma} + D_{\varphi} \right)$.

Similar with trRosetta, we have developed our in-house inter-residue distance and orientation predictor. Our key improvements over trRosetta include:
\begin{itemize}
\item \textbf{Network architecture.} We adopt a much deeper residual network (over 600+ layers), trained under a progressive learning manner \cite{karras2017progressive, bachlechner2020rezero}. Criss-cross attention \cite{huang2019ccnet} and squeeze-and-excitation block \cite{hu2018squeeze} are adopted better capture the global information for inter-residue distance and orientation prediction.
\item \textbf{Multi-database fusion.} We search over multiple protein sequence databases (UniClust30 \cite{mirdita2016uniclust}, UniRef90 \cite{suzek2014uniref}, etc.) with various hyper-parameter combinations (E-value and number of iterations) to obtain multiple groups of MSA data. Independent predictions are made from each MSA data, and then ranked to filtered out low-confidence ones, clustered to group similar predictions, and then averaged to produce final predictions.
\end{itemize}

We entered the CASP14 contact prediction competition with this inter-residue distance predictor, and ranked $1^{\text{st}}$ among all the 60 teams. Therefore, we use this predictor to produce high-quality predictions, which is one of the critical input features for the subsequent structure optimization module. It is worth mentioning that our \ours{} framework is general and can also take other inter-residue distance and/or orientation predictors' outputs as input features.

\section{\ours{}}

In this section, we describe technical details of our proposed \ours{} approach. To start with, we discuss how protein structures are converted into the score network's inputs. Afterwards, we introduce the training process of \ours{}'s underlying score network, and how it can be employed to enable fully-differentiable protein structure optimization via sampling. Finally, we conclude this section with detailed discussions on several critical implementation details of our \ours{} approach.

\subsection{From 3D Coordinates to 2D Distance Matrix}

The original protein structure optimization task requires determining all the atoms' 3D coordinates in the protein, based on the given amino-acid sequence. This problem can often be simplified to solving 3D coordinates for backbone atoms ($N$, $C_{\alpha}$, and $C'$) or even $C_{\alpha}$ atoms only. This is because that once the backbone or $C_{\alpha}$-trace is solved, then the side-chain conformation can determined by various off-the-shelf methods, \eg{} SCWRL4 \cite{krivov2009improved}. Here, we concentrate on how to accurately solve the $C_{\alpha}$-trace in a fully-differentiable manner.

For a protein with $L$ residues, we denote all the $C_{\alpha}$ atoms' 3D coordinates as $\mathbf{X} \in \mathbb{R}^{L \times 3}$. To enable the denoising score matching training, we perturb the ground-truth 3D coordinate matrix $\mathbf{X}$ (based on experimental structures) with random noise drawn from the isotropic Gaussian distribution $\mathcal{N} \left( \mathbf{0}, \sigma^{2} \mathbf{I} \right)$. We denote the perturbed 3D coordinate matrix as $\tilde{\mathbf{X}}$, which follows the conditional probability distribution $q_{\sigma} ( \tilde{\mathbf{X}} | \mathbf{X} )$.

One may directly train a score network that takes the perturbed 3D coordinate matrix $\tilde{\mathbf{X}}$ as inputs, and then attempt to predict the log probability's gradients over $\tilde{\mathbf{X}}$, given by:
\begin{equation}
\nabla_{\tilde{\mathbf{X}}} \log q_{\sigma} ( \tilde{\mathbf{X}} | \mathbf{X} ) = - \frac{\tilde{\mathbf{X}} - \mathbf{X}}{\sigma^{2}}
\end{equation}
which is the difference between perturbed and ground-truth 3D coordinates, scaled by random noise's standard deviation.

However, there can be infinite number of equivalent 3D coordinate matrices to represent the same protein structure, due to arbitrary 3D rotations\footnotemark{}. This requires the score network to be SO(3)-equivariant, \ie{} if input 3D coordinates are rotated by some 3D rotation matrix, then the score network's outputs should be equivalently rotated, by the same rotation matrix. SE(3)-Transformer \cite{fuchs2020se3} and LieTransformer \cite{hutchinson2021lietransformer} are qualified for such requirements, and can be used as an alternative choice for the architecture of \ours{}'s score network.

\footnotetext{For the same structure, 3D translations also lead to infinite number of equivalent 3D coordinate matrices. However, this can be easily resolved by subtracting the averaged 3D coordinate from each atom's 3D coordinate.}

Here, we propose another approach to bypass the above SO(3)-equivariance requirement. Let $\mathbf{D} \in \mathbb{R}^{L \times L}$ denote the squared Euclidean distance matrix, \ie{} $d_{ij} = \left\| \mathbf{x}_{i} - \mathbf{x}_{j} \right\|_{2}^{2}$ where $\mathbf{x}_{i}$ is the 3D coordinate of the $i$-th residue's $C_{\alpha}$ atom. For a given protein structure, its corresponding distance matrix is unique; on the other hand, a valid distance matrix\footnotemark{} has two corresponding protein structures, only differ by the mirroring operation. The correct protein structure and its mirrored counterpart can be distinguished by considering the handedness, and shall be discussed in Section \ref{sec:impl_details}. Therefore, we can bypass the SO(3)-equivariance requirement for score networks with distance matrix based representations for protein structures.

\footnotetext{We call a distance matrix valid, if there exists at least one corresponding 3D coordinate matrix.}

\subsection{Training}

\begin{figure}
\includegraphics[width=\linewidth]{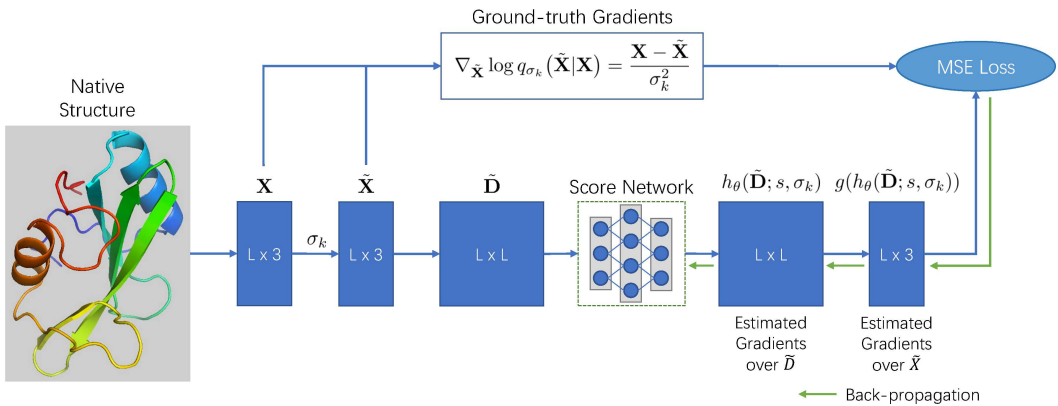}
\caption{The overall workflow of \ours{}'s training process. Given a native protein structure, we extract all the $C_{\alpha}$ atoms' 3D coordinates $\mathbf{X}$, and impose random noise to obtain perturbed 3D coordinates $\tilde{\mathbf{X}}$. It is then converted into a distance matrix to be fed into the score network to predict the corresponding gradients, which are further transformed into estimated gradients over perturbed 3D coordinates. The estimated and ground-truth gradients are then compared under the MSE loss function to provide training signals for the score network.}
\label{fig:workflow_training}
\end{figure}

Prior to structure optimization via sampling in \ours{}, we need to train a score network to produce estimations of log probability's gradients over distance matrices. The score network is trained in a similar manner as that in \cite{song2019generative}. The basic idea is to perturb native protein structures with various levels of random noise, and then let the score network to estimate true gradients over perturbed distance matrices. The overall workflow of \ours{}'s training process is as depicted in Figure \ref{fig:workflow_training}.

Let $z$ be an amino-acid sequence and $\mathbf{X}$ be its native protein structure, represented by all the residues' $C_{\alpha}$ atoms' 3D coordinates. The perturbation of 3D coordinates is accomplished by the isotropic Gaussian random noise, \ie{} the random noise is independently added to each atom's X/Y/Z-axis coordinate. We choose a series of random noise's standard deviations $\left\{ \sigma_{k} \right\}$, where $\sigma_{1} > \sigma_{2} > \dots > \sigma_{K}$ and $K$ is the number of random noise levels. For the $k$-th level, the perturbed data distribution is defined as:
\begin{equation}
q_{\sigma_{k}} \big( \tilde{\mathbf{X}} | \mathbf{X} \big) = \mathcal{N} \left( \mathbf{X} | \sigma_{k}^{2} \mathbf{I} \right) = \frac{1}{\left( 2 \pi \sigma_{k}^{2} \right)^{3L / 2}} \cdot \exp \left( - \frac{1}{2 \sigma_{k}^{2}} \left\| \tilde{\mathbf{X}} - \mathbf{X} \right\|_{F}^{2} \right)
\end{equation}
and its log probability's gradients over perturbed 3D coordinates $\tilde{\mathbf{X}}$ are given by:
\begin{equation}
\nabla_{\tilde{\mathbf{X}}} \log q_{\sigma_{k}} \big( \tilde{\mathbf{X}} | \mathbf{X} \big) = \frac{\mathbf{X} - \tilde{\mathbf{X}}}{\sigma_{k}^{2}}
\label{eqn:ldfold_log_prob_grad_v2}
\end{equation}
which will serve as the supervised information for training the score network.

We denote the score network as $h_{\theta} ( \tilde{\mathbf{D}}; s, \sigma_{k} )$, where the distance matrix $\tilde{\mathbf{D}}$ is computed from perturbed 3D coordinates $\tilde{\mathbf{X}}$, and the amino-acid sequence $s$ and random noise's standard deviation $\sigma_{k}$ constitute conditional inputs. The score network follows a standard 2D convolutional network architecture, consists of multiple residual blocks but without any pooling layers to keep a consistent spatial size of feature maps. The score network's outputs, denoted as $\mathbf{H} \in \mathbb{R}^{L \times L}$, aim at approximating the log probability's gradients over the perturbed distance matrix.

However, since the random noise is not directly imposed on the distance matrix, it is non-trivial to obtain ground-truth gradients over the perturbed distance matrix. Hence, we use the score network's estimation of distance matrix's gradients to derive its estimation of perturbed 3D coordinates' gradients. Based on the chain rule, we have:
\begin{equation}
\mathbf{g}_{i} = \sum\nolimits_{j = 1}^{L} 2 \left( h_{ij} + h_{ji} \right) \left( \mathbf{x}_{i} - \mathbf{x}_{j} \right)
\label{eqn:chain_rule}
\end{equation}
where $\mathbf{g}_{i}$ denotes the estimated gradients for the $i$-th residue $C_{\alpha}$ atom's 3D coordinate vector $\mathbf{x}_{i}$. We stack all the $\left\{ \mathbf{g}_{i} \right\}$ to form the overall estimation of gradients over perturbed 3D coordinates, denoted as $\mathbf{G}$. We use $g \left( \cdot \right)$ to denote the mapping from $\mathbf{H}$ to $\mathbf{G}$, thus we have $\mathbf{G} = g ( h_{\theta} ( \tilde{\mathbf{D}}; s, \sigma_{k} ) )$.

The optimization objective function for the score network is then given by:
\begin{equation}
\frac{1}{2 N K} \sum_{\mathbf{X} \in \mathcal{X}} \sum_{k = 1}^{K} \sigma_{k}^{2} \cdot \mathbb{E}_{\tilde{\mathbf{X}} \sim \mathcal{N} \left( \mathbf{X} | \sigma_{k}^{2} \mathbf{I} \right)} \left\| g ( h_{\theta} ( \tilde{\mathbf{D}}; s, \sigma_{k} ) ) - \frac{\mathbf{X} - \tilde{\mathbf{X}}}{\sigma_{k}^{2}} \right\|_{F}^{2}
\label{eqn:ldfold_obj_func}
\end{equation}
where $\mathcal{X}$ is the set of all the native protein structures, and $N = \left| \mathcal{X} \right|$ is its cardinality. Please note that $\tilde{\mathbf{D}}$ is computed from perturbed 3D coordinates $\tilde{\mathbf{X}}$, but we omit this relationship here for simplicity. By minimizing the above loss function, we force the score network to simultaneously approximate the perturbed data distributions' log probability's gradients over perturbed 3D coordinates for all the random noise levels. Once trained, the score network is then used in the upcoming sampling process for structure optimization.

\subsection{Sampling}

\begin{figure}
\includegraphics[width=\linewidth]{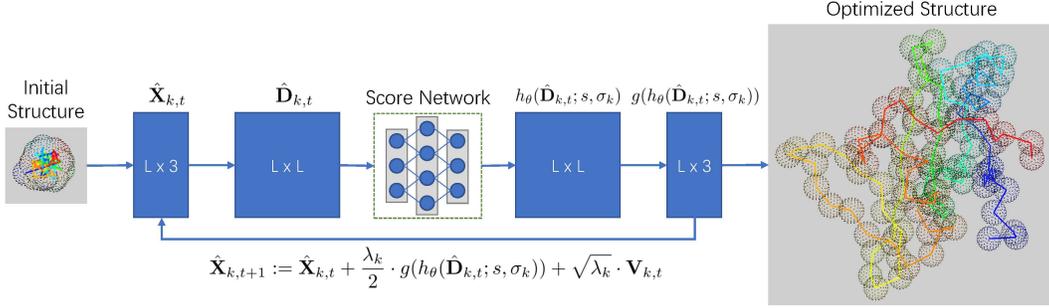}
\caption{The overall workflow of \ours{}'s sampling process.}
\label{fig:workflow_sampling}
\end{figure}

Similar with \cite{song2019generative}, we adopt the annealed Langevin dynamics sampling to gradually optimize protein structures from random initialization. The key difference is that we need to ensure the validness of distance matrices throughout the sampling process, \ie{} the distance matrix should always correspond to some valid structure in the 3D space. Therefore, instead directly using the score network's estimated gradients to update its inputs, which may break the distance matrix's validness, we update 3D coordinates with its estimated gradients derived via the chain rule. We illustrate the overall workflow of this sampling process in Figure \ref{fig:workflow_sampling} and Algorithm \ref{alg:ldfold_sampling}.

The annealed Langevin dynamics sampling is divided into multiple stages, one per random noise level as used in the training process. This allows more radical exploration at early stages to find more promising initial poses, and then gradually reduces the step size for more fine-grained structure optimization. It is worth mentioning that our \ours{} approach does not require an initial protein structure to start with. Instead, we simply initialize 3D coordinates for all the $C_{\alpha}$ atoms from the prior distribution $\mathcal{N} \left( \mathbf{0}, \mathbf{I} \right)$.

We denote the number of iterations within each stage as $T$, and the protein structure's 3D coordinates at the $k$-th stage's $t$-th iteration is denoted as $\hat{\mathbf{X}}_{k, t}$. The pairwise distance matrix $\hat{\mathbf{D}}_{k, t}$ is computed and fed into the score network, obtaining its estimated gradients. Following Eq. (\ref{eqn:chain_rule}), we derive the estimated gradients over 3D coordinates, and update the structure via:
\begin{equation}
\hat{\mathbf{X}}_{k, t + 1} := \hat{\mathbf{X}}_{k, t + 1} + \frac{\lambda_{k}}{2} \cdot g ( h_{\theta} ( \hat{\mathbf{D}}_{k, t}; s, \sigma_{k} ) ) + \sqrt{\lambda_{k}} \cdot \mathbf{V}_{k, t}
\label{eqn:ldfold_update}
\end{equation}
where $\lambda_{k}$ is the step size for the $k$-th stage, and $\mathbf{V}_{k, t} \in \mathbb{R}^{L \times 3}$ consists of random noise drawn from the normal distribution $\mathcal{N} \left( 0, 1 \right)$. The step size is determined based on the corresponding random noise's standard deviation, given by:
\begin{equation}
\lambda_{k} = \lambda_{0} \sigma_{k}^{2}
\label{eqn:ldfold_stepsize}
\end{equation}
which is exactly the same scheme as used in \cite{song2019generative}. This approximately guarantees a constant signal-to-noise ratio throughout the annealed Langevin dynamics sampling process, regardless of the random noise level. Each stage's terminal structure is used as the next stage's initialization, \ie{} $\hat{\mathbf{X}}_{k + 1, 0} = \hat{\mathbf{X}}_{k, T}$.

\begin{algorithm}
\begin{algorithmic}[1]
\REQUIRE score network $h_{\theta} ( \cdot )$, number of iterations per stage $T$, reference step size $\lambda_{0}$
\ENSURE optimized 3D coordinates $\hat{\mathbf{X}}_{K, T}$
\STATE initialize $\hat{\mathbf{X}}_{1, 0}$ from $\mathcal{N} \left( \mathbf{0}, \mathbf{I} \right)$
\FOR{$k = 1, \dots, K$}
\STATE compute the step size $\lambda_{k} = \lambda_{0} \sigma_{k}^{2}$
\FOR{$t = 0, \dots, T - 1$}
\STATE compute the distance matrix $\hat{\mathbf{D}}_{k, t}$ from $\hat{\mathbf{X}}_{k, t}$
\STATE compute $g ( h_{\theta} ( \hat{\mathbf{D}}_{k, t}; s, \sigma_{k} ) )$ (estimated gradients over 3D coordinates)
\STATE sample random noise $\mathbf{V}_{k, t}$ from $\mathcal{N} \left( 0, 1 \right)$
\STATE update 3D coordinates $\hat{\mathbf{X}}_{k, t + 1}$ by Eq. (\ref{eqn:ldfold_update})
\ENDFOR
\ENDFOR
\end{algorithmic}
\caption{\ours{} Sampling}
\label{alg:ldfold_sampling}
\end{algorithm}

\subsection{Implementation Details}
\label{sec:impl_details}

Below, we present implementation details of our \ours{} approach's training and sampling procedures, including: 1) how to build up a score network, 2) how to construct protein-specific conditional inputs to the score network, and 3) how the handedness issue can be resolved.

\begin{figure}
\centering
\includegraphics[width=.8\linewidth]{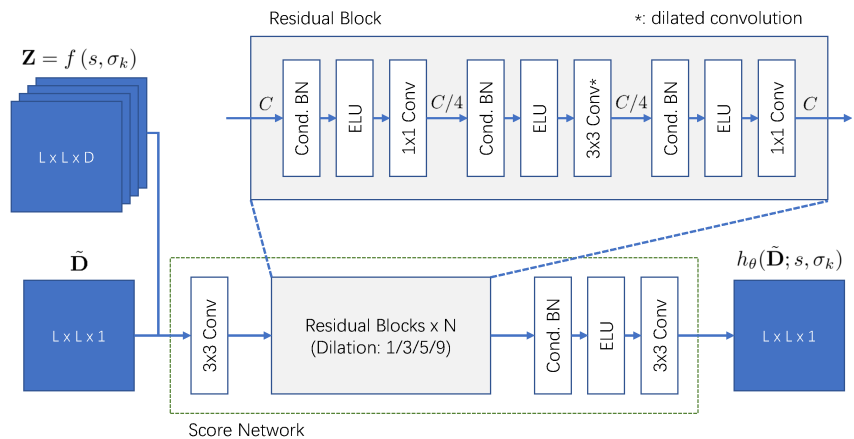}
\caption{The score network's architecture.}
\label{fig:network_arch}
\end{figure}

\textbf{Network architecture.} Since protein structures are encoded as 2D distance matrices, we can adopt any fully-convolutional neural networks as the backbone architecture of the score network. Concretely, we build a score network with residual units as its basic building blocks. Each residual unit uses the bottleneck mechanism to reduce the computational overhead, and employs conditional batch normalization \cite{song2019generative} to take random noise's standard deviation level into consideration. We visualize the score network's architecture in Figure \ref{fig:network_arch}.

The perturbed distance matrix $\tilde{\mathbf{D}}$, together with conditional inputs $\mathbf{Z} = f \left( s, \sigma_{k} \right)$, are fed into the score network. We shall discuss how conditional inputs are constructed in the next section. Multiple residual blocks are used, whose dilation rates are iteratively selected from $\left\{ 1, 3, 5, 9 \right\}$. The dilated convolution allows a larger receptive field with the same computational complexity, and widely used for inter-residue distance and orientation predictions \cite{senior2019protein, senior2020improved, yang2020improved}. After passing through the final output block (Cond. BN + ELU + Conv), the score network outputs its estimation for the log probability's gradients over the perturbed distance matrix.

\textbf{Protein-specific conditional inputs.} We encode the protein-specific information into score network's input features from three aspects: 1) amino-acid sequence's one-hot encoding; 2) positional encoding; and 3) inter-residue distance and orientation predictions.

\textbf{1) One-hot encoding:} For an amino-acid sequence of length $L$, we encode it into a $L \times 20$ matrix of one-hot encoding vectors, and then repeat it row-wisely and column-wisely to be stacked into a 2D feature map of size $L \times L \times 40$.

\textbf{2) Positional encoding:} In order to encode each residue's relative position in the sequence, we adopt a similar positional encoding scheme as in \cite{vaswani2017attention}. Assume we want the positional encoding to form a feature map of size $L \times L \times D_{PE}$, we firstly encode the sequence into a $L \times \frac{1}{2} D_{PE}$ matrix as below:
\begin{equation}
\begin{split}
Z_{i, 2 r} &= \sin \left[ \sfrac{i}{\text{pow} ( L_{max}, \frac{4r}{D_{PE}} )} \right] \\
Z_{i, 2 r + 1} &= \cos \left[ \sfrac{i}{\text{pow} ( L_{max}, \frac{4r}{D_{PE}} )} \right]
\end{split}
\end{equation}
where $r = 0, 1, \dots, \lfloor \frac{1}{4} D_{PE} \rfloor$ and $L_{max} = 1000$ is selected as the ``maximal'' sequence length, although this also works for amino-acid sequences longer than $L_{max}$. Afterwards, we repeat this matrix row-wisely and column-wisely to form a 2D feature map, similar as above.

\textbf{3) Distance and orientation predictions:} Our in-house inter-residue distance and orientation predictor uses the same prediction format as trRosetta. Such predictions are naturally packed as 2D feature maps, where the number of channels is given by $D_{d} = 37$, $D_{\omega} = D_{\gamma} = 25$, and $D_{\varphi} = 13$.

We stack all these features along the channel dimension to form protein-specific conditional inputs.

\textbf{The handedness issue.} As mentioned earlier, a valid distance matrix corresponds to two protein structures, one being the other one's mirrored counterpart. To distinguish the one with correct handedness, we rely on the distribution of dihedral angles defined on adjacent residues' $C_{\alpha}$ atoms.

\begin{figure}
\centering
\hspace*{\fill}
\begin{subfigure}{.3\textwidth}
  \centering
  \includegraphics[width=\linewidth]{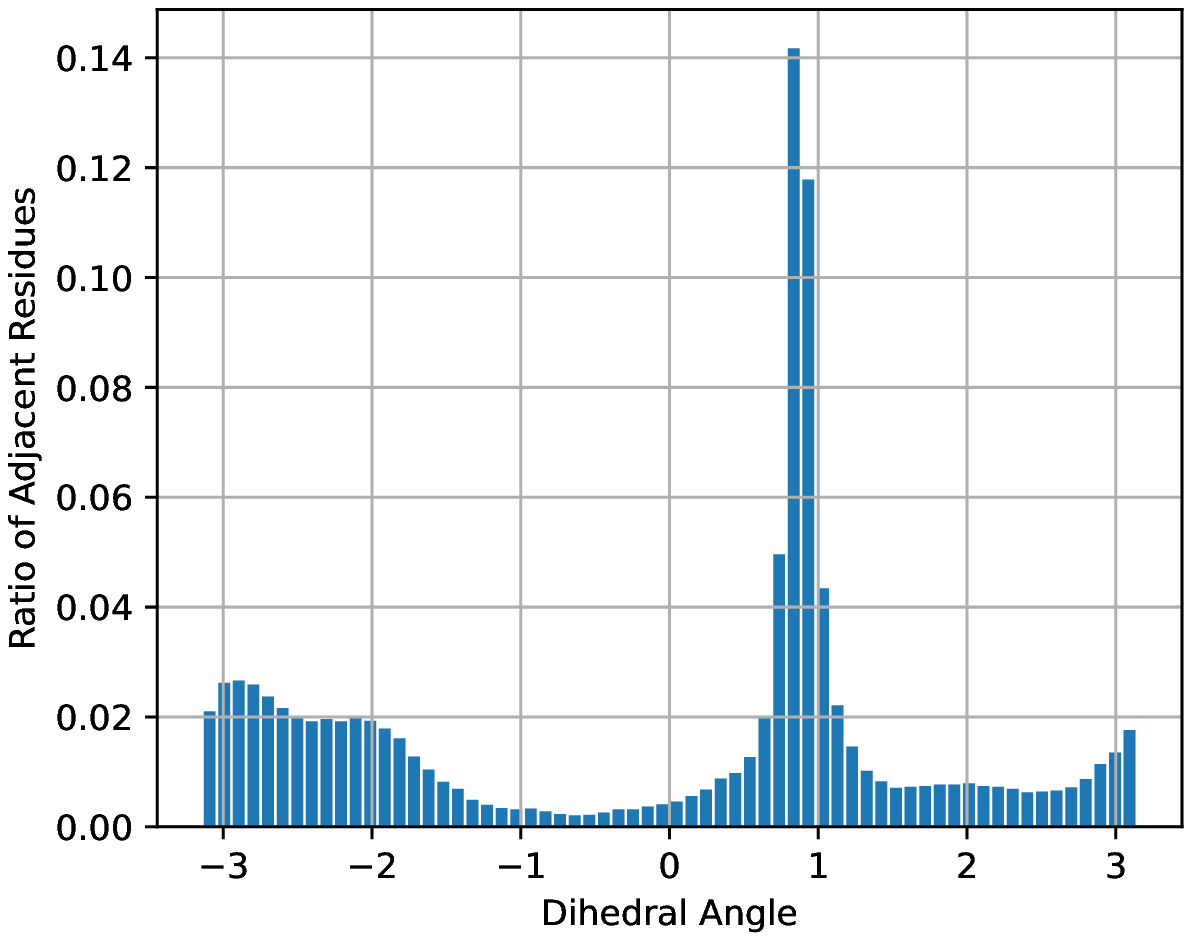}
  \caption{Reference}
  \label{fig:ca_dihd_stats_ref}
\end{subfigure}
\begin{subfigure}{.3\textwidth}
  \centering
  \includegraphics[width=\linewidth]{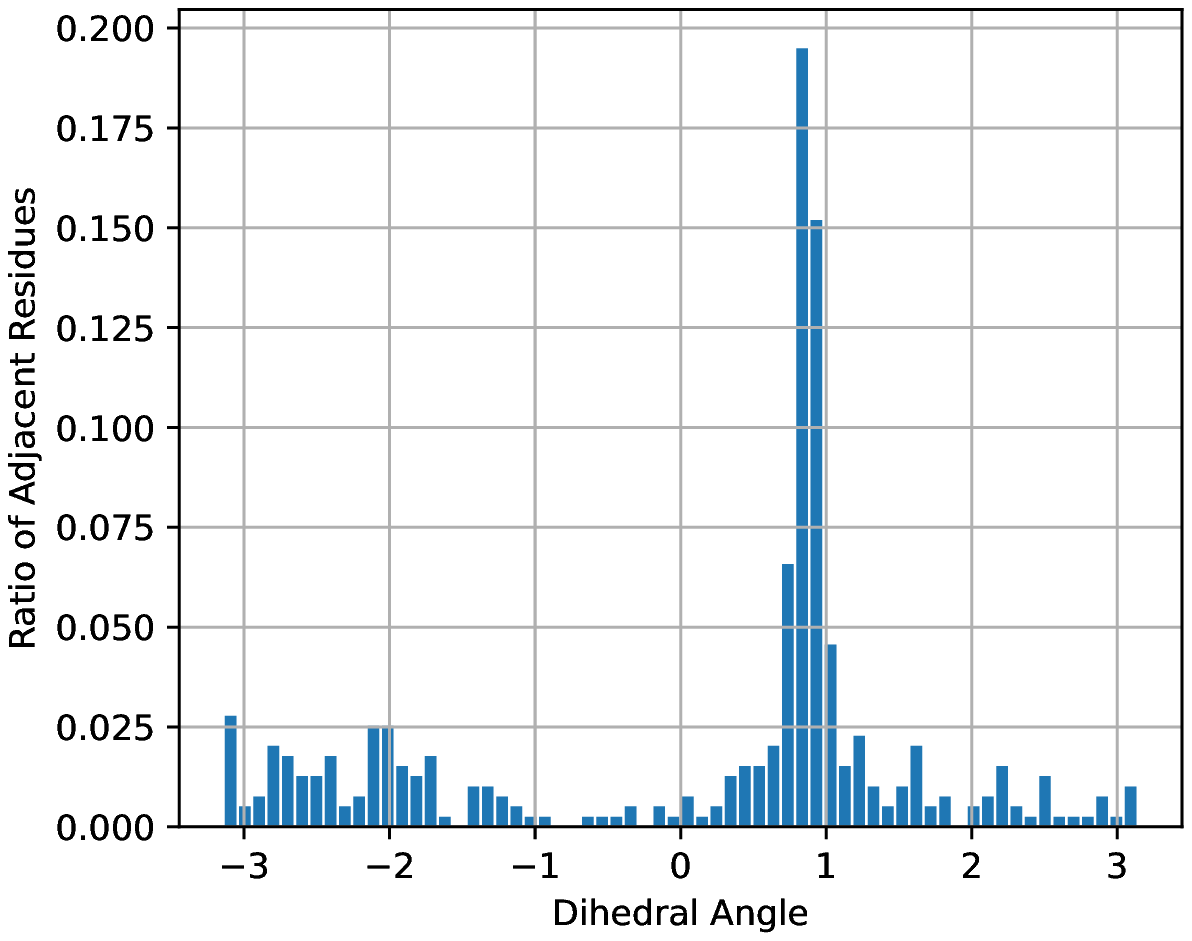}
  \caption{1JFB-A00 (Original)}
  \label{fig:ca_dihd_stats_raw}
\end{subfigure}
\begin{subfigure}{.3\textwidth}
  \centering
  \includegraphics[width=\linewidth]{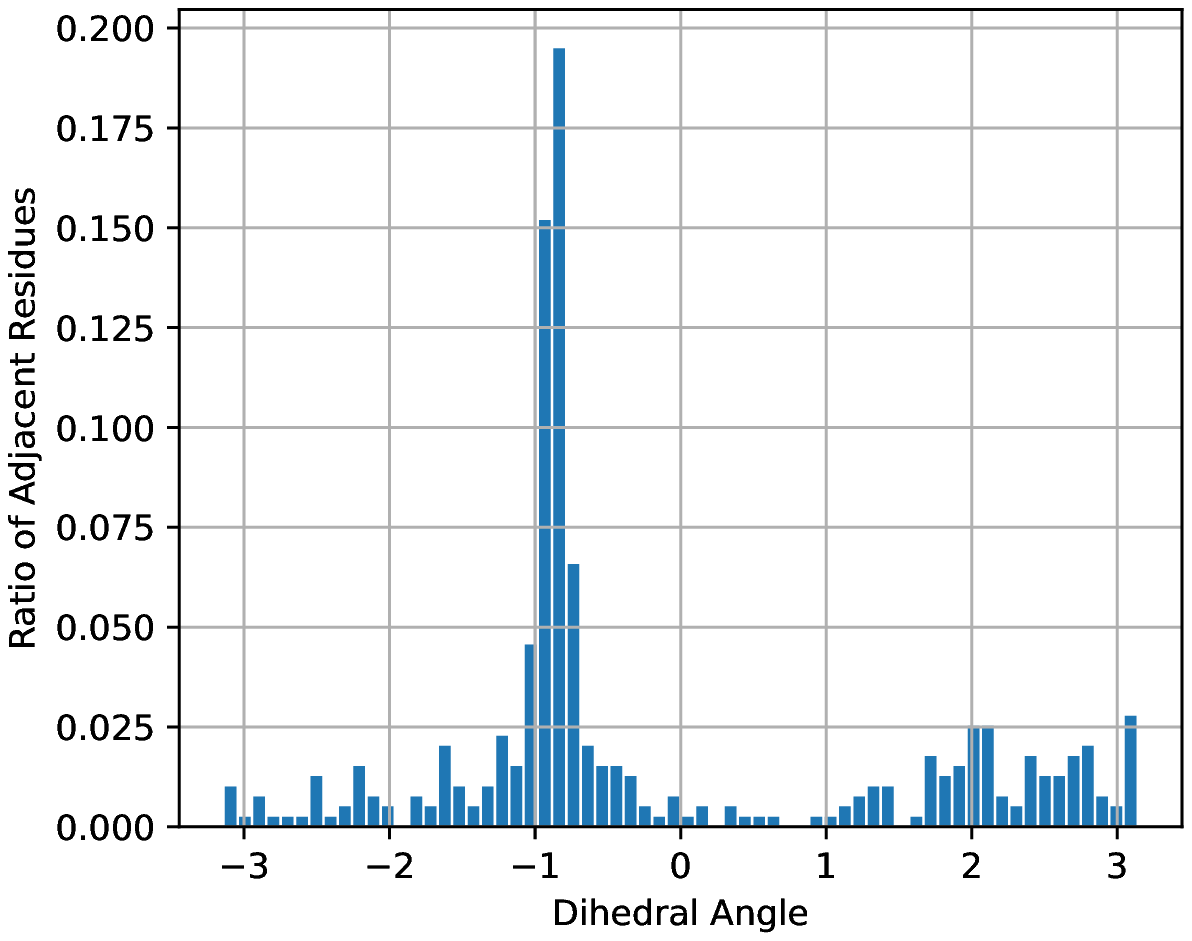}
  \caption{1JFB-A00 (Mirrored)}
  \label{fig:ca_dihd_stats_mrr}
\end{subfigure}
\hspace*{\fill}
\caption{Comparison on dihedral angle distributions, defined on adjacent residues' $C_{\alpha}$ atoms.}
\label{fig:ca_dihd_stats}
\end{figure}

In Figure \ref{fig:ca_dihd_stats_ref}, we plot the distribution of dihedral angles of adjacent residues' $C_{\alpha}$ atoms, computed on the CATH-Train subset (see Section \ref{sec:experiments} for details on the dataset). As for comparison, we randomly select one domain from the dataset (ID: 1JFB-A00), and plot its original and mirrored structures' dihedral angle distributions, as depicted in Figure \ref{fig:ca_dihd_stats_raw} and \ref{fig:ca_dihd_stats_mrr}.

It is obvious that the original one is more similar with the reference distribution. For quantitative analysis, we compute the KL-divergence \cite{kullback1951information} for original and mirrored structures against the reference distribution. The original structure's KL-divergence is merely 0.1353, much lower than the mirrored one's 1.9919. Therefore, we propose the handedness issue resolving module (HIRM) to utilize this characteristic. Specifically, at the end of each annealed Langevin dynamics sampling stage (Algorithm \ref{alg:ldfold_sampling}), we compute the dihedral angle distribution's KL-divergence for each structure and its mirrored counterpart, and select the one with lower KL-divergence. This procedure can fix the handedness issue with negligible computational overhead, as we shall demonstrate in the experiments.

\section{Experiments}
\label{sec:experiments}

To verify the effectiveness and efficiency of our proposed \ours{} approach, we conduct extensive experiments on protein structure optimization, and compare its performance against one of the state-of-the-art structure optimization pipelines, trRosetta \cite{yang2020improved}. Specifically, we aim at answering the following two questions:

\begin{itemize}
\item Can \ours{} generate high-quality decoys for structure optimization?
\item Can \ours{} efficiently optimize structures from random initialization?
\end{itemize}

\subsection{Setup}

To train and evaluate our \ours{} approach, we construct a medium-scale domain-level dataset, following the domain definition in the CATH database \cite{sillitoe2020cath}. We use CATH's daily snapshot (Oct 21st, 2020) and randomly select 2065 domains, split into 1665/200/200 as training/validation/test subsets. It is guaranteed that these three subsets are disjoint in the super-family level, \ie{} each CATH super-family only appears in at most one subset.

Our \ours{}'s score network consists of 32 residual blocks with dilation convolution, and the number of channels in hidden layers' feature maps is set to 64. We use a batch size of 64 for training and validation, and apply random cropping of size 32 to input feature maps for data augmentation. The dimension of positional encodings is set to $D_{PE} = 48$. We construct $K = 32$ levels of random noise's standard deviations, ranging from 0.01 to 10.0, as $\sigma_{1} = 10.0$ can sufficiently explore the conformation space while $\sigma_{K} = 0.01$ only introduces negligible perturbation to native structures. The score network is trained with an Adam optimizer \cite{kingma2015adam} for 200 epochs, with a constant learning rate $0.0001$. Afterwards, we select the optimal checkpoint based on the validation loss, and then use it for the upcoming structure optimization.

For structure optimization, both trRosetta and \ours{} require inter-residue distance and orientation predictions as prerequisite inputs. To ensure a fair comparison, we adopt our in-house inter-residue distance and orientation predictor to generate such predictions, and feed them into both structure optimization pipelines. For each test target, we generate 300 decoys with trRosetta, modeled with 15 different hyper-parameter combinations\footnotemark{} and 20 decoys per combination. As for our proposed \ours{} approach, we discover that the decoy quality variance is much lower than that of trRosetta (to be discussed later); therefore, we only generate 128 decoys per test target for evaluation. The number of iterations within each \ours{}'s structure optimization stage is set to $T = 64$, and the reference step size $\lambda_{0} = 0.1$.

\footnotetext{Following the original trRosetta's implementation, we use following hyper-parameter combinations: 1) PCUT thresholds: 0.05, 0.15, 0.25, 0.35, and 0.45; and 2) optimization modes: S+M+L, SM+L, and SML (S: short; M: medium; L: long).}

We use lDDT-Ca \cite{mariani2013lddt} as the evaluation metric (higher the better), which measure the local distance differences of $C_{\alpha}$ atoms between the native protein structure and predicted structure decoys.

In terms of hardware, we use one Nvidia V100 GPU card, equipped with an Intel Xeon Platinum 8255C CPU, for both \ours{}'s training and structure optimization. It takes around 18 hours to train \ours{}'s score network under such hardware specification.

\subsection{Decoy Quality}

\begin{figure}
\centering
\hspace*{\fill}
\begin{subfigure}{.4\textwidth}
  \centering
  \includegraphics[width=\linewidth]{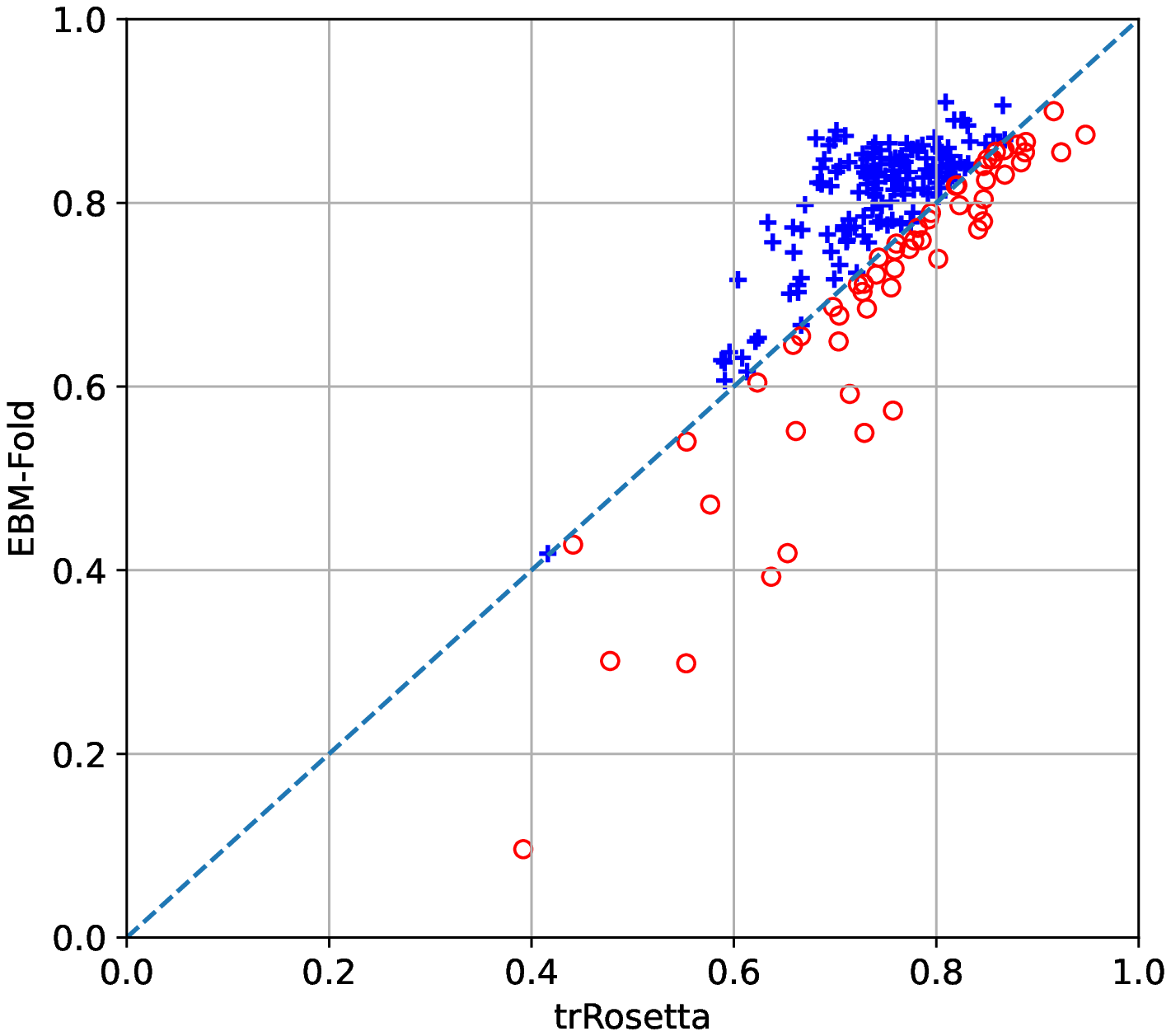}
  \caption{lDDT-Ca-Avg}
  \label{fig:lddt_scat_avg_wr}
\end{subfigure}
\hfill
\begin{subfigure}{.4\textwidth}
  \centering
  \includegraphics[width=\linewidth]{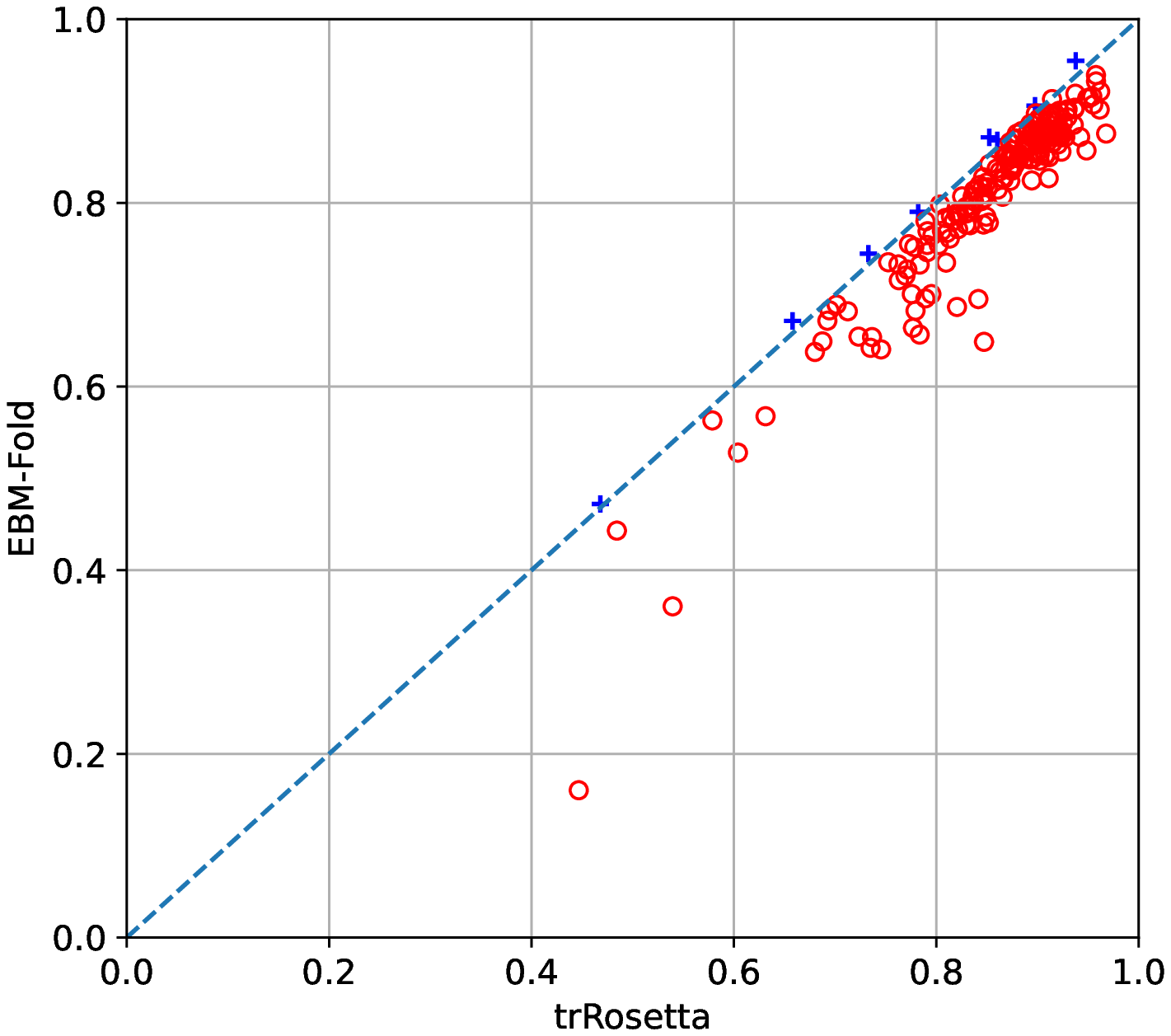}
  \caption{lDDT-Ca-Max}
  \label{fig:lddt_scat_max_wr}
\end{subfigure}
\hspace*{\fill}
\caption{Comparison on trRosetta and \ours{}'s per-target decoy quality, with both inter-residue distance and orientation predictions provided as inputs.}
\label{fig:lddt_scat_wr}
\end{figure}

To start with, we compare trRosetta and \ours{}'s ability in generating high-quality decoys, given the same inter-residue distance and orientation predictions as inputs. In Figure \ref{fig:lddt_scat_wr}, we present the per-target comparison between trRosetta and \ours{}'s averaged/maximal lDDT-Ca scores. The averaged results across all the 200 test targets are listed in Table \ref{tab:lddt_comp}.

\begin{table}
\centering
\caption{Comparison on trRosetta and \ours{}'s averaged decoy quality, with either distance+orientation or distance-only predictions provided as inputs.}
\begin{tabular}{c|c|c|c|c}
\hline
\multirow{2}{*}{Method} & \multicolumn{2}{c|}{Distance + Orientation} & \multicolumn{2}{c}{Distance Only} \\ \cline{2-5}
 & lDDT-Ca-Avg & lDDT-Ca-Max & lDDT-Ca-Avg & lDDT-Ca-Max \\ \hline \hline
trRosetta \cite{yang2020improved} & 0.7452 & \textbf{0.8523} & 0.4394 & 0.8017 \\ \hline
\ours{} & \textbf{0.7775} & 0.8124 & \textbf{0.7708} & \textbf{0.8043} \\ \hline
\end{tabular}
\label{tab:lddt_comp}
\end{table}

\begin{figure}
\centering
\hspace*{\fill}
\begin{subfigure}{.3\textwidth}
  \centering
  \includegraphics[width=\linewidth]{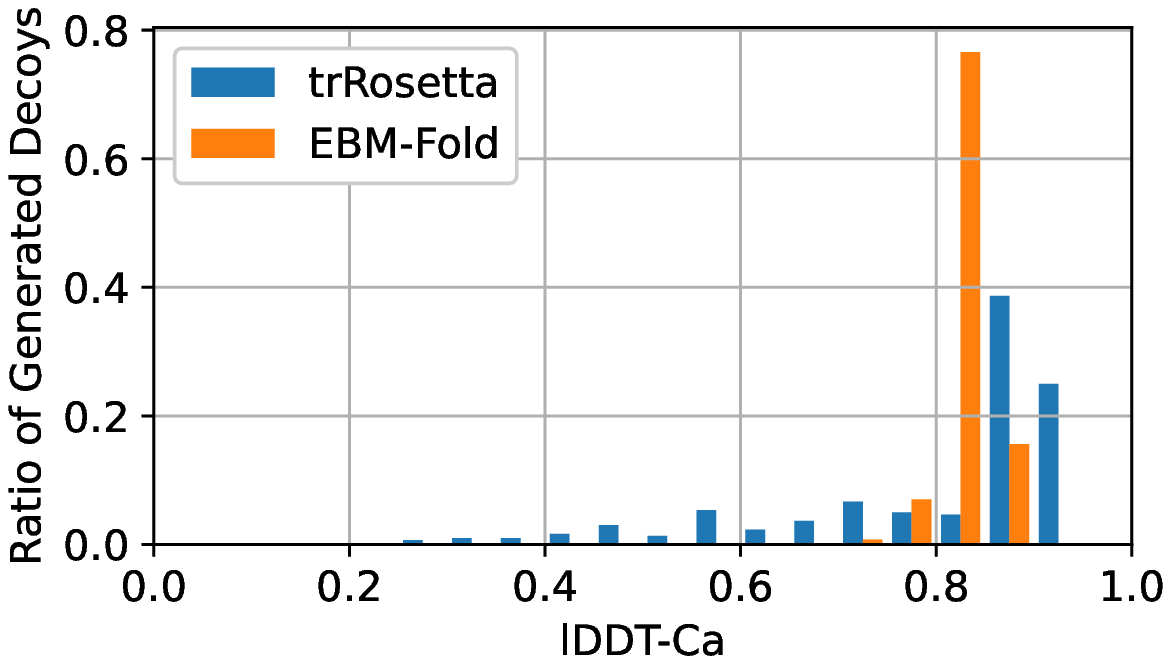}
  \caption{1AHO-A00 ($L = 64$)}
\end{subfigure}
\begin{subfigure}{.3\textwidth}
  \centering
  \includegraphics[width=\linewidth]{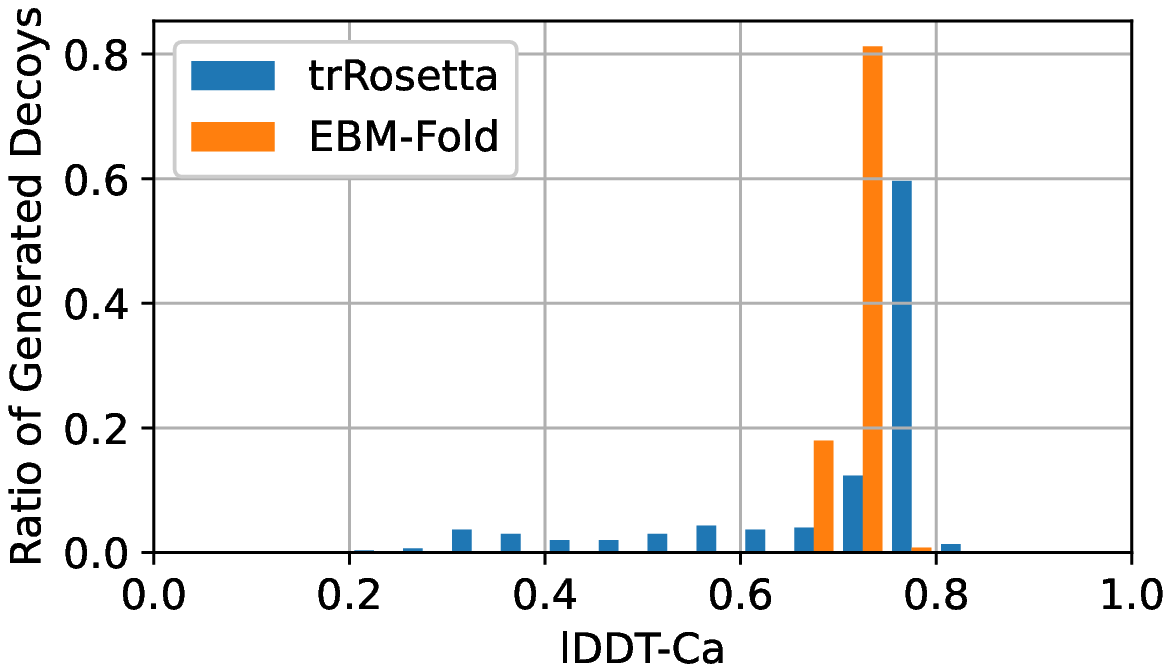}
  \caption{2BSJ-A00($L = 128$)}
\end{subfigure}
\begin{subfigure}{.3\textwidth}
  \centering
  \includegraphics[width=\linewidth]{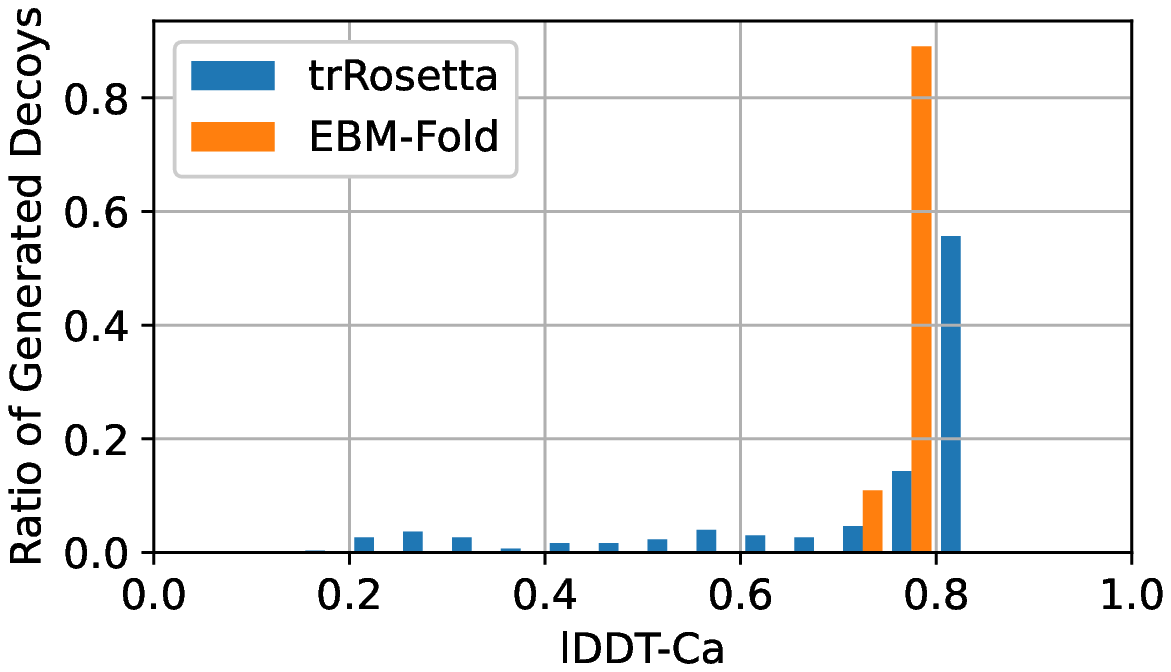}
  \caption{3A2Z-A00($L = 190$)}
\end{subfigure}
\hspace*{\fill}
\caption{Comparison on trRosetta and \ours{}'s per-target lDDT-Ca distributions, with both inter-residue distance and orientation predictions provided as inputs.}
\label{fig:lddt_hist_wr}
\end{figure}

From Figure \ref{fig:lddt_scat_wr} and Table \ref{tab:lddt_comp}, we discover that when both inter-residue distance and orientation predictions are used, our \ours{} approach is inferior to trRosetta, based on the comparison on maximal lDDT-Ca scores (Figure \ref{fig:lddt_scat_max_wr}, 0.8124 vs. 0.8523). However, the averaged lDDT-Ca scores of \ours{} is higher than that of trRosetta (Figure \ref{fig:lddt_scat_avg_wr}, 0.7775 vs. 0.7452), suggesting that \ours{}'s decoy quality distribution is more concentrated, while trRosetta has a larger variance in per-target decoys' lDDT-Ca scores. This is further verified by the visualization of per-target lDDT-Ca distributions, as depicted in Figure \ref{fig:lddt_hist_wr}.

From Figure \ref{fig:lddt_hist_wr}, we observe that trRosetta's lDDT-Ca scores span a much larger range, while \ours{}'s decoy quality is more likely to distributed within two or three adjacent bins (bin width: 0.05). It is worth further investigation on how to improve the \ours{}'s top-ranked decoy quality.

\begin{figure}
\centering
\hspace*{\fill}
\begin{subfigure}{.4\textwidth}
  \centering
  \includegraphics[width=\linewidth]{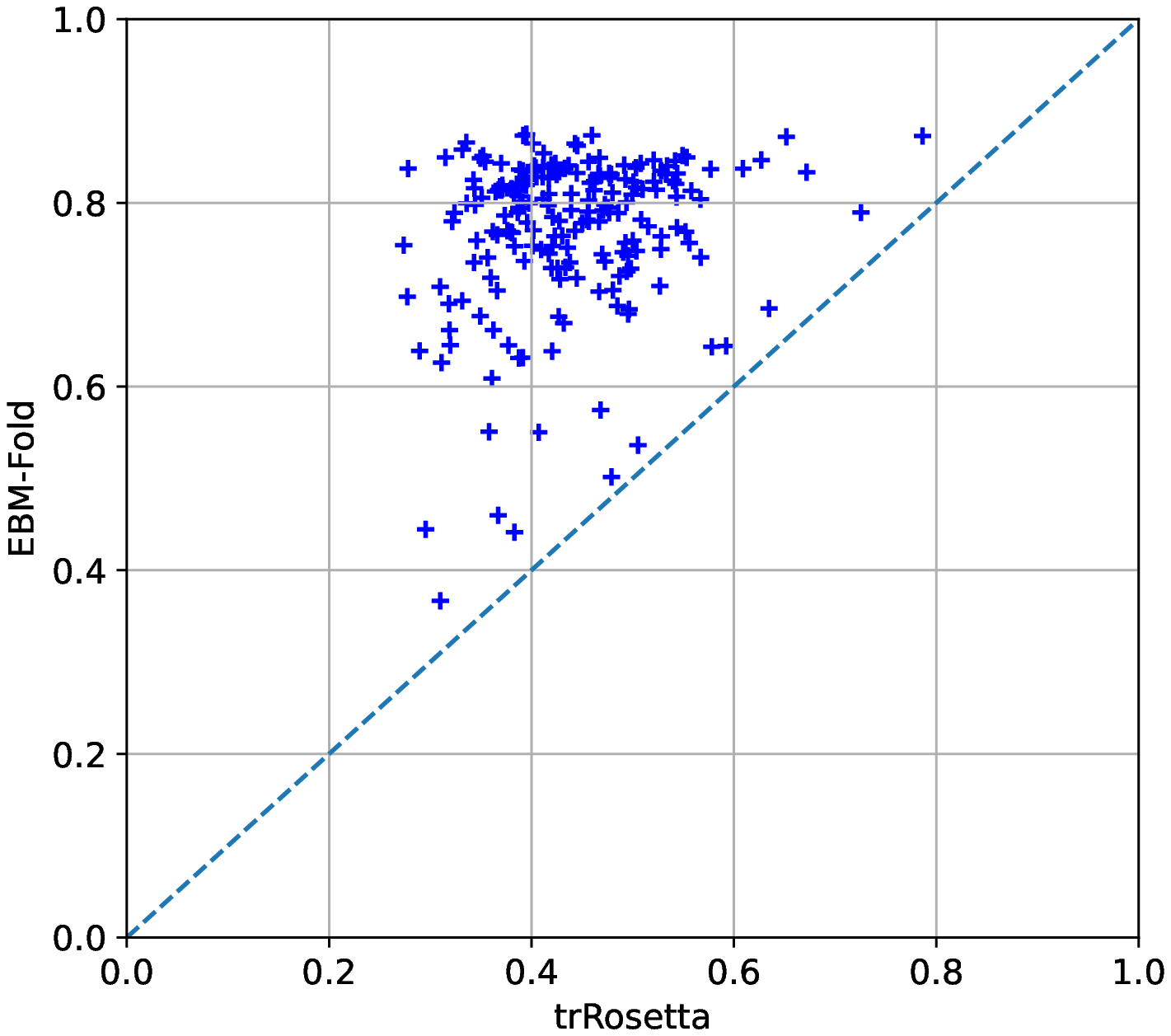}
  \caption{lDDT-Ca-Avg}
  \label{fig:lddt_scat_avg_wo}
\end{subfigure}
\hfill
\begin{subfigure}{.4\textwidth}
  \centering
  \includegraphics[width=\linewidth]{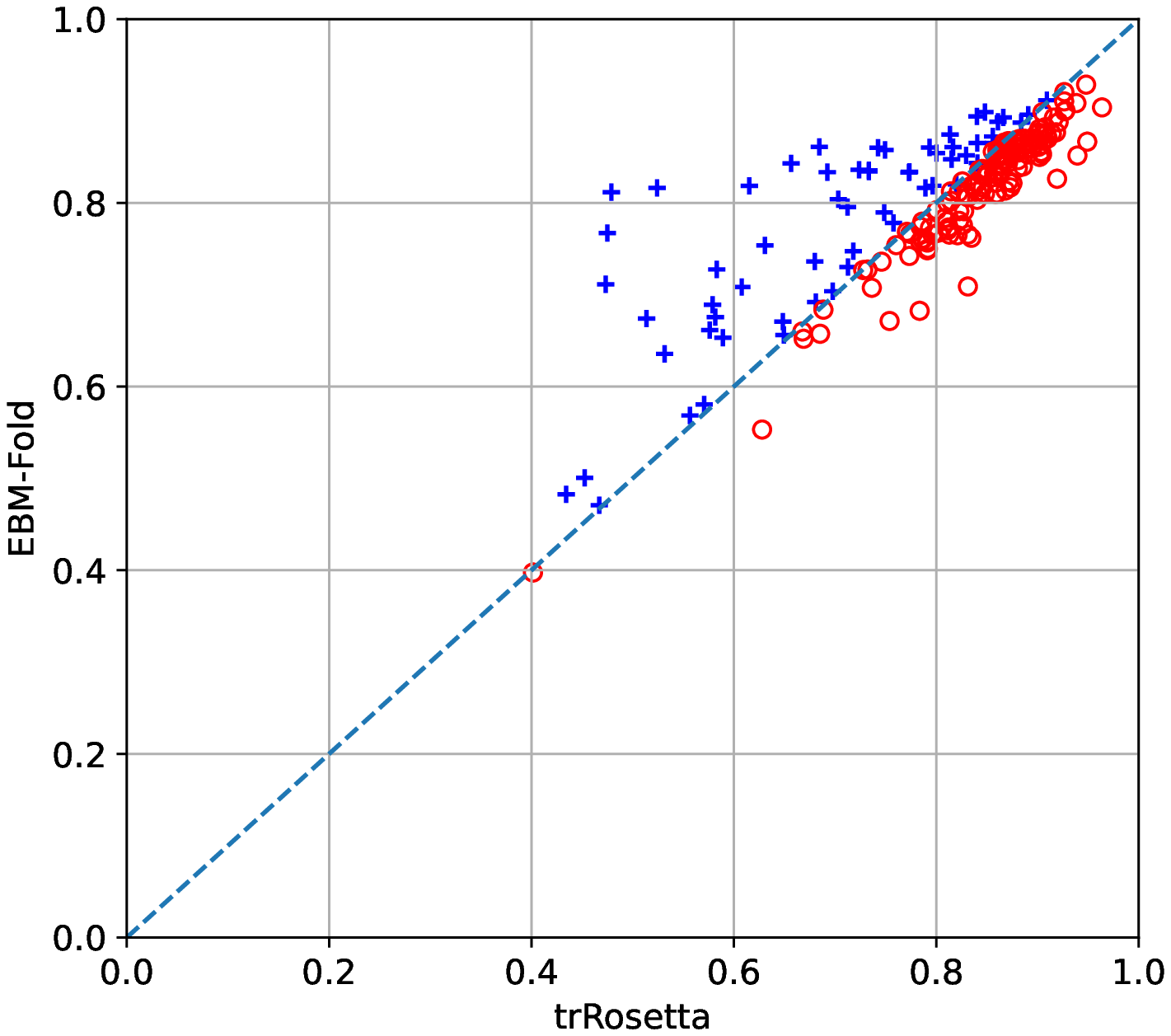}
  \caption{lDDT-Ca-Max}
  \label{fig:lddt_scat_max_wo}
\end{subfigure}
\hspace*{\fill}
\caption{Comparison on trRosetta and \ours{}'s per-target decoy quality, with only inter-residue distance predictions provided as inputs.}
\label{fig:lddt_scat_wo}
\end{figure}

\begin{figure}
\centering
\hspace*{\fill}
\begin{subfigure}{.3\textwidth}
  \centering
  \includegraphics[width=\linewidth]{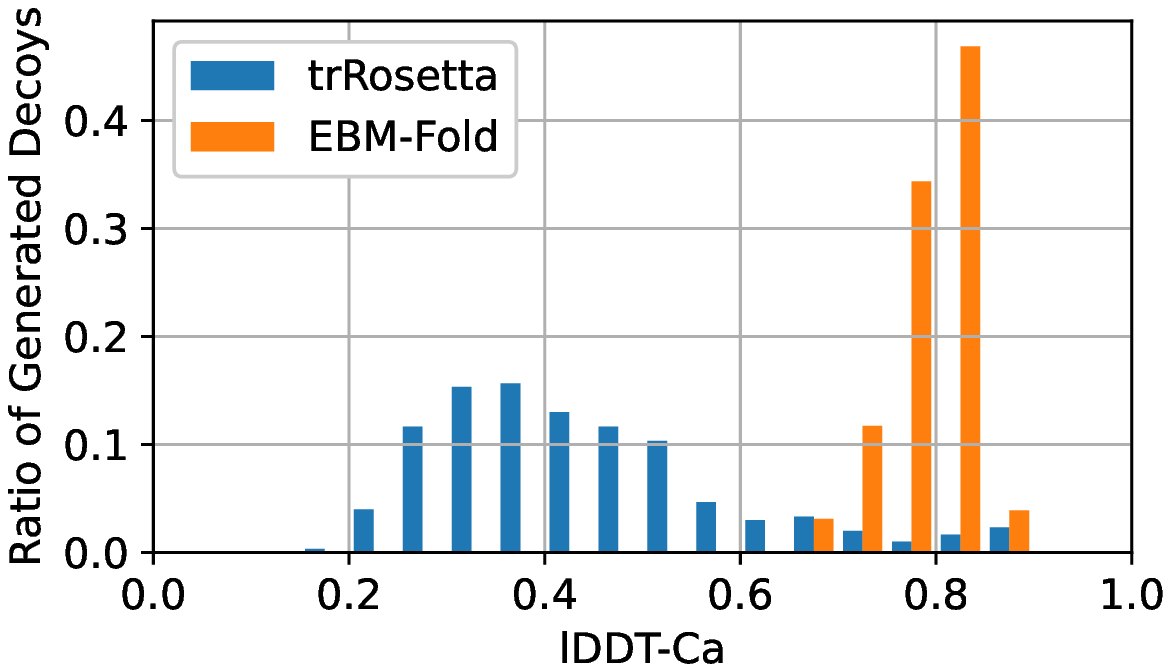}
  \caption{1AHO-A00 ($L = 64$)}
\end{subfigure}
\begin{subfigure}{.3\textwidth}
  \centering
  \includegraphics[width=\linewidth]{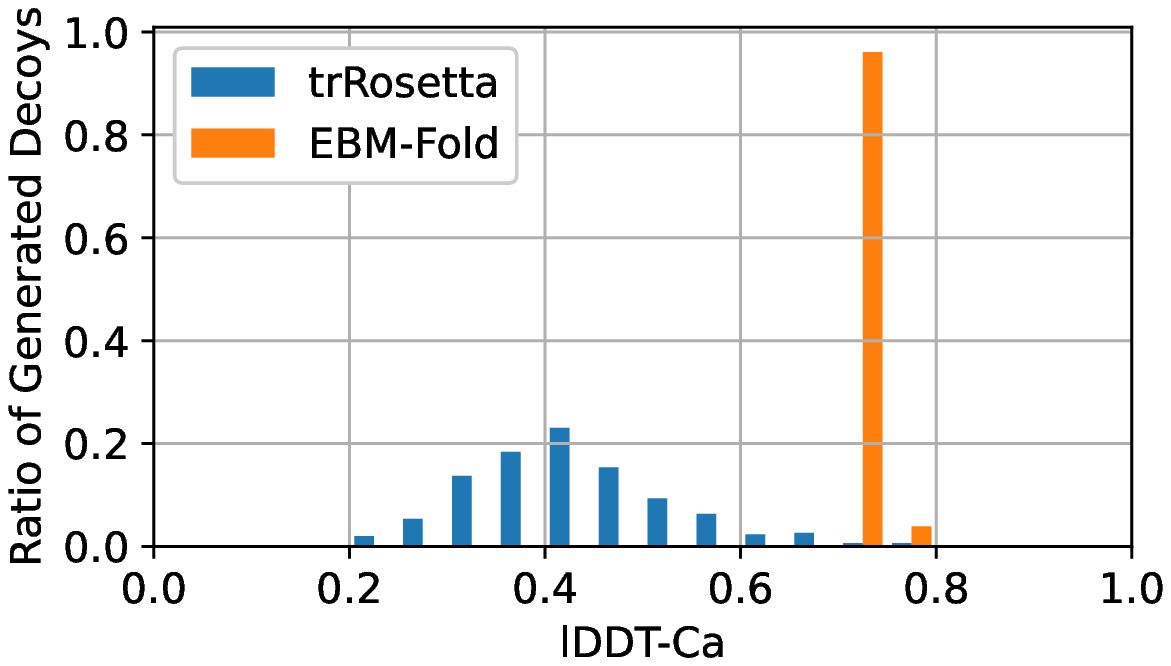}
  \caption{2BSJ-A00 ($L = 128$)}
\end{subfigure}
\begin{subfigure}{.3\textwidth}
  \centering
  \includegraphics[width=\linewidth]{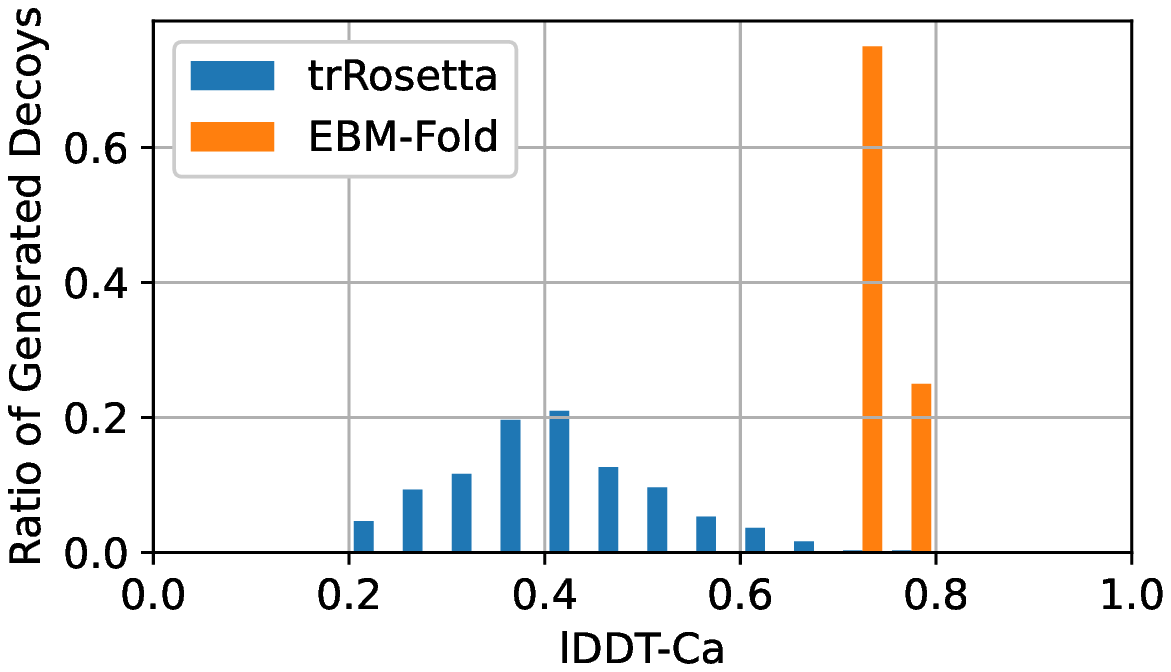}
  \caption{3A2Z-A00 ($L = 190$)}
\end{subfigure}
\hspace*{\fill}
\caption{Comparison on trRosetta and \ours{}'s per-target lDDT-Ca distributions, with only inter-residue distance predictions provided as inputs.}
\label{fig:lddt_hist_wo}
\end{figure}

Unlike trRosetta, \ours{} does not explicitly use inter-residue distance and orientation predictions as restraints for structure optimization. Instead, \ours{} takes both predictions as the score network's conditional inputs. Since both predictions are derived from the same inter-residue distance and orientation predictor, it is very likely that they are highly correlated. Therefore, we conduct an additional experiments to only use inter-residue distance predictions for structure optimization. The corresponding per-target and averaged results are presented in Figure \ref{fig:lddt_scat_wo} and Table \ref{tab:lddt_comp}. Similarly, we visualize both methods' lDDT-Ca distributions in Figure \ref{fig:lddt_hist_wo}.

It is observed that trRosetta suffers a significant drop in lDDT-Ca scores, indicating that inter-residue orientation predictions are indeed critical in improving the structure optimization performance, as emphasized in the original paper \cite{yang2020improved}. However, the performance degradation of \ours{} is much smaller than that of trRosetta, suggesting that due to the high correlation between these two predictions, \ours{} is more robust to the removal of inter-residue orientation prediction. This also points out a possible direction for improving \ours{}'s decoy quality: how to explicitly cooperate distance and orientation predictions, rather than simply using them as input features to the score network.

\subsection{Optimization Efficiency}

The computational efficiency is also one of major concerns for structure optimization methods. Our \ours{} approach relies on the annealed Langevin dynamics sampling process to gradually optimize structures from random initialization. As stated earlier, each iteration only requires one forward pass with the score network, and no gradient computation is needed. This process can be fully implemented on GPUs, while most of traditional structure optimization methods are more CPU-extensive.

\begin{figure}
\centering
\includegraphics[width=.8\linewidth]{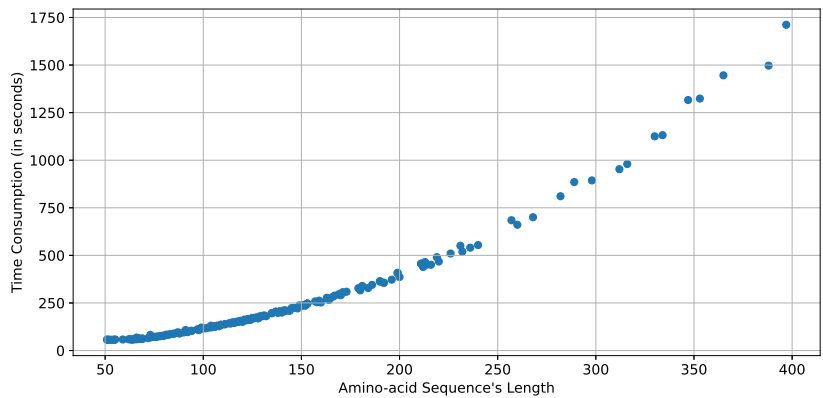}
\caption{Visualization of \ours{}'s structure optimization process's time consumption under various amino-acid sequence lengths.}
\label{fig:runtime}
\end{figure}

In Figure \ref{fig:runtime}, we report \ours{}'s structure optimization process's time consumption for proteins with various sequence length. As GPU is more efficient for a larger batch size, we set the batch size to 16 and results reported here are the overall time for all the 16 different initial structures being simultaneously optimized. Since each protein is encoded as a $L \times L$ feature map as inputs, the time complexity of each forward pass is quadratic to the sequence length $L$. This is verified in Figure \ref{fig:runtime}, as the time consumption grows quadratically to the sequence length. For proteins with fewer than 200 amino-acids, the overall time consumption for generating 16 optimized structures is under 8 minutes, indicating that \ours{}'s structure optimization process is quite efficient.

\subsection{Visualization}

\begin{figure}
\centering
\hspace*{\fill}
\begin{subfigure}{.3\textwidth}
  \centering
  \includegraphics[width=\linewidth]{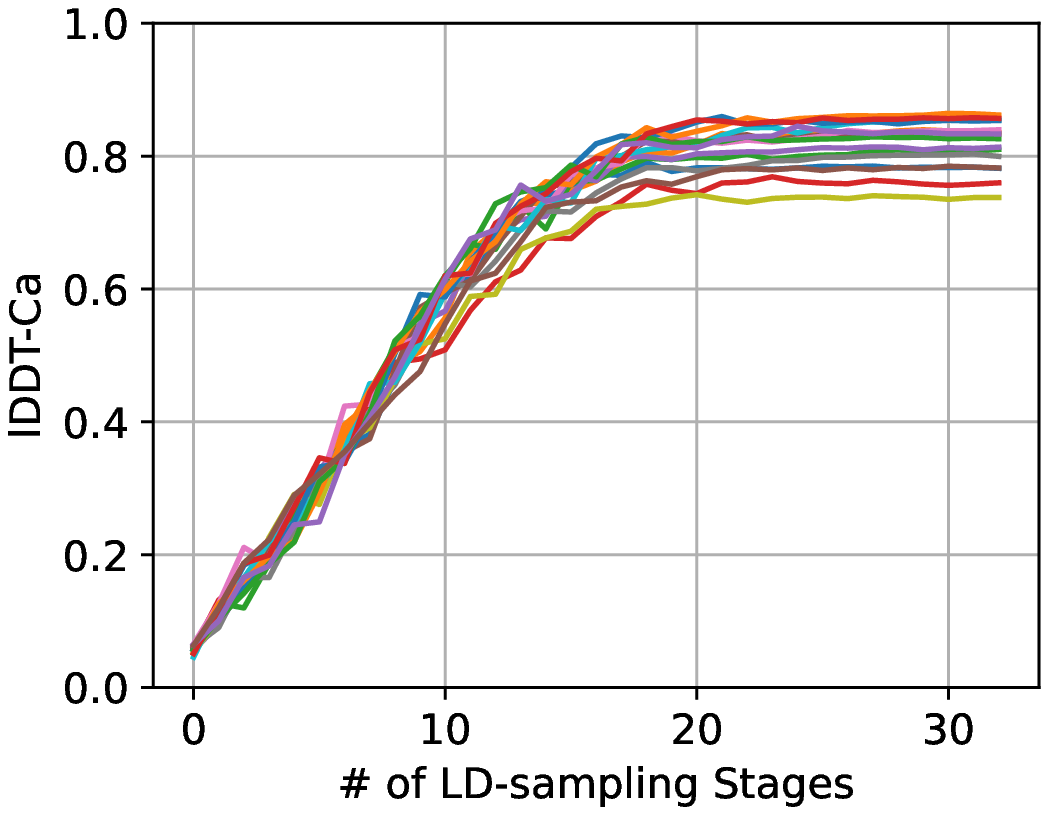}
  \caption{1AHO-A00 ($L = 64$)}
\end{subfigure}
\begin{subfigure}{.3\textwidth}
  \centering
  \includegraphics[width=\linewidth]{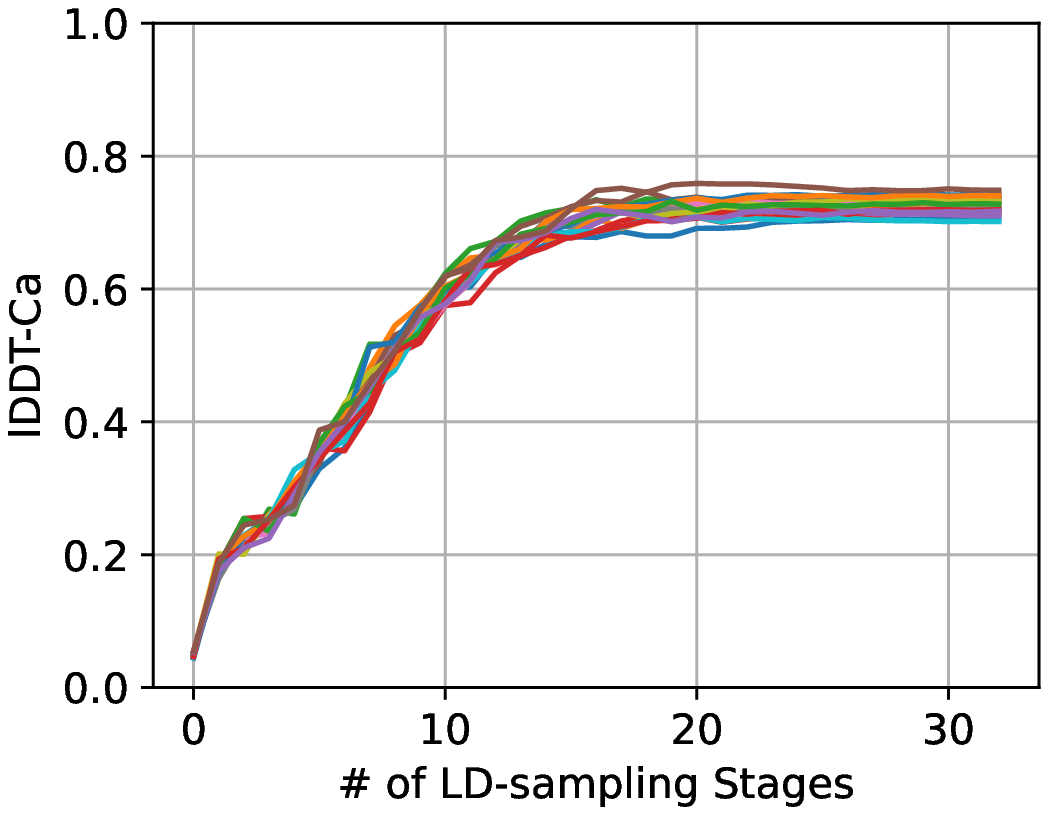}
  \caption{2BSJ-A00 ($L = 128$)}
\end{subfigure}
\begin{subfigure}{.3\textwidth}
  \centering
  \includegraphics[width=\linewidth]{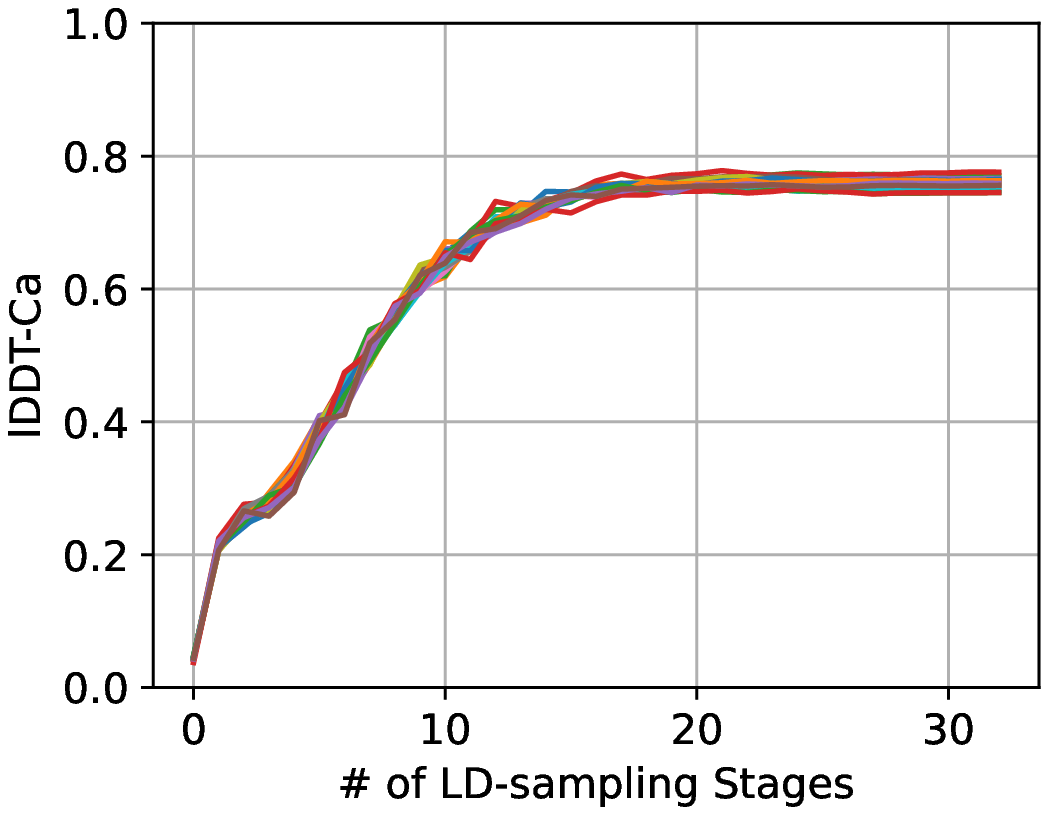}
  \caption{3A2Z-A00 ($L = 190$)}
\end{subfigure}
\hspace*{\fill}
\caption{Visualization of \ours{}'s structure optimization process, measured by lDDT-Ca scores at the end of each LD-sampling stage.}
\label{fig:lddt_curve}
\end{figure}

Below, we present detailed visualization of \ours{}'s structure optimization process, to see how a randomly initialized structure is gradually optimized. In Figure \ref{fig:lddt_curve}, we plot the lDDT-Ca versus the number of annealed Langevin dynamics sampling stages curve for three selected targets.

From Figure \ref{fig:lddt_curve}, we observe that although initial structures' lDDT-Ca scores are low (under 0.1), \ours{}'s structure optimization leads a consistent improvement in lDDT-Ca scores, especially during the first $\sfrac{2}{3}$ LD-sampling stages. This corresponds to the coarse-to-fine optimization with a gradually reduced step size, as the random noise's standard deviation $\sigma_{k}$ decreases from $10.0$ to $0.1$, and the step size $\lambda_{k}$ is quadratic to $\sigma_{k}$, as defined in Eq. (\ref{eqn:ldfold_stepsize}). The remaining LD-sampling stages' step size  is even smaller; therefore, the improvement in lDDT-Ca scores is less significant.

\begin{figure}
\centering
\begin{subfigure}{\textwidth}
  \includegraphics[width=\linewidth]{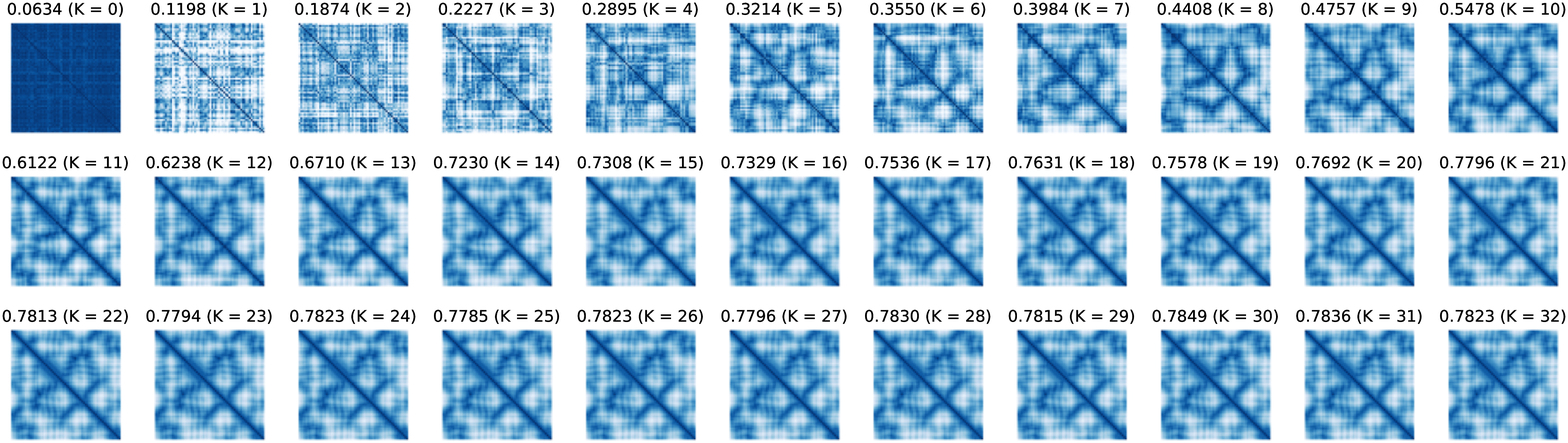}
  \caption{1AHO-A00 ($L = 64$)}
\end{subfigure}
\begin{subfigure}{\textwidth}
  \includegraphics[width=\linewidth]{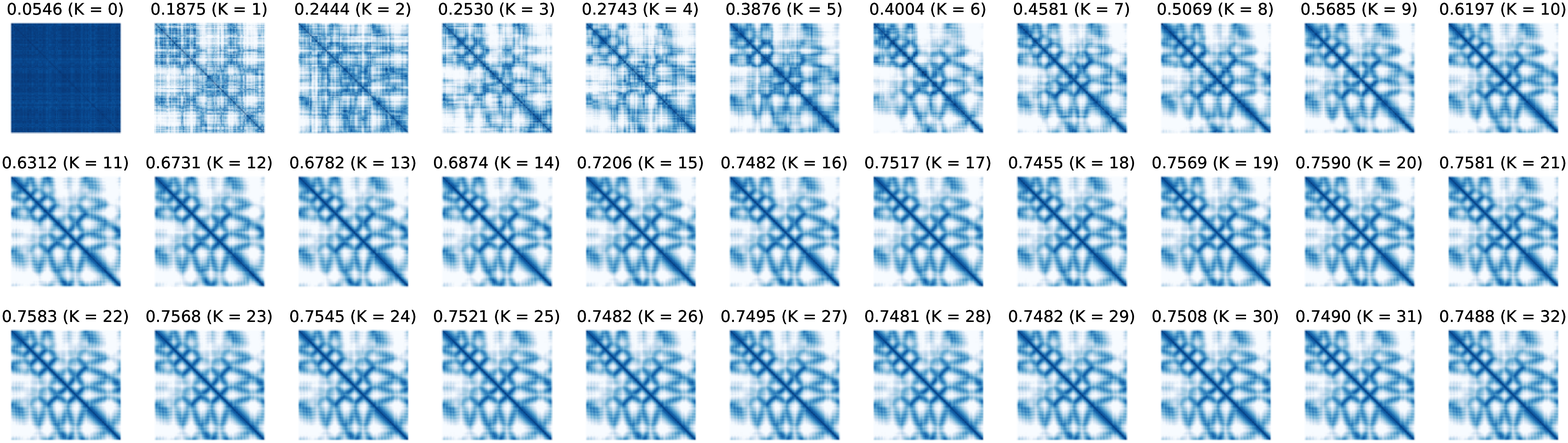}
  \caption{2BSJ-A00 ($L = 128$)}
\end{subfigure}
\begin{subfigure}{\textwidth}
  \includegraphics[width=\linewidth]{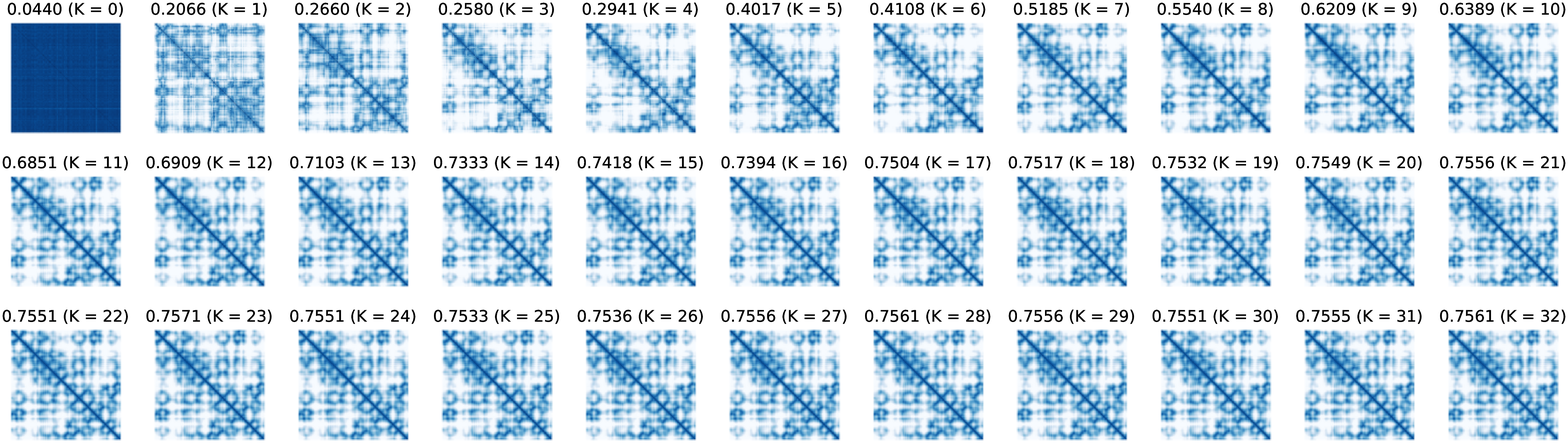}
  \caption{3A2Z-A00 ($L = 190$)}
\end{subfigure}
\caption{Visualization of distance matrices' dynamics in \ours{}'s structure optimization process. The lDDT-Ca score and number of LD-sampling stages elapsed are noted in each sub-figure.}
\label{fig:lddt_dist_map}
\end{figure}

To dive deeper into \ours{}'s structure optimization process, we further visualize intermediate structures' distance matrices in Figure \ref{fig:lddt_dist_map}. Since initial structures' 3D coordinates are simply drawn from $\mathcal{N} \left( 0, 1 \right)$, all the pairwise distance values are small, as illustrated in each sub-figure top-left element (deeper color corresponds to smaller distance). As \ours{}'s structure optimization goes on, we clearly observe that major patterns in the distance matrix gradually emerge and lDDT-Ca scores stably improve.

\subsection{Ablation Study}

\begin{figure}
\centering
\hspace*{\fill}
\begin{subfigure}{.24\textwidth}
  \includegraphics[width=\linewidth]{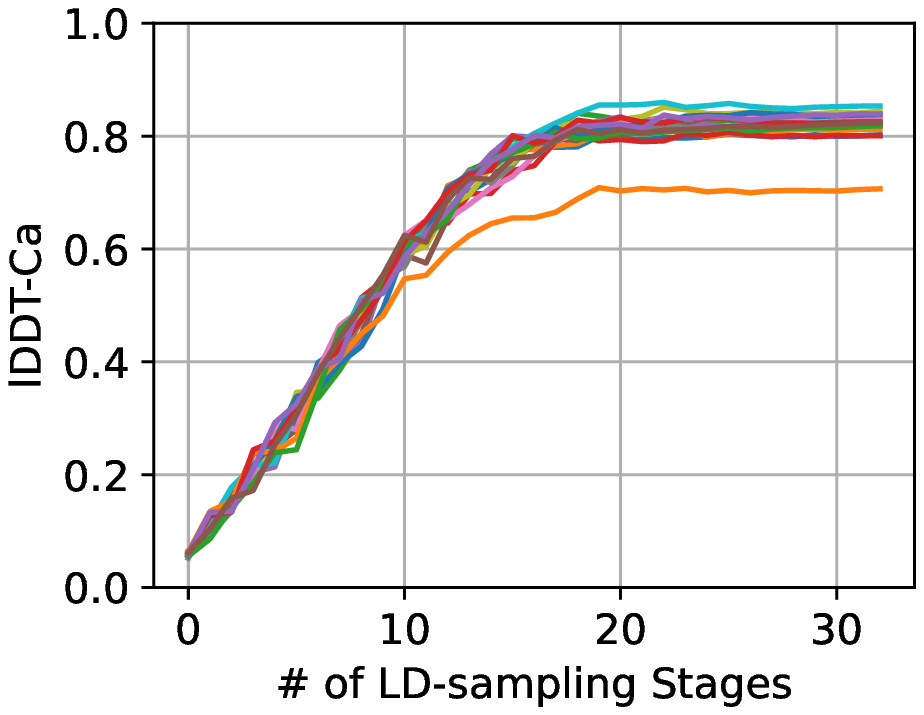}
  \caption{lDDT-Ca w/o HIRM}
\end{subfigure}
\begin{subfigure}{.24\textwidth}
  \includegraphics[width=\linewidth]{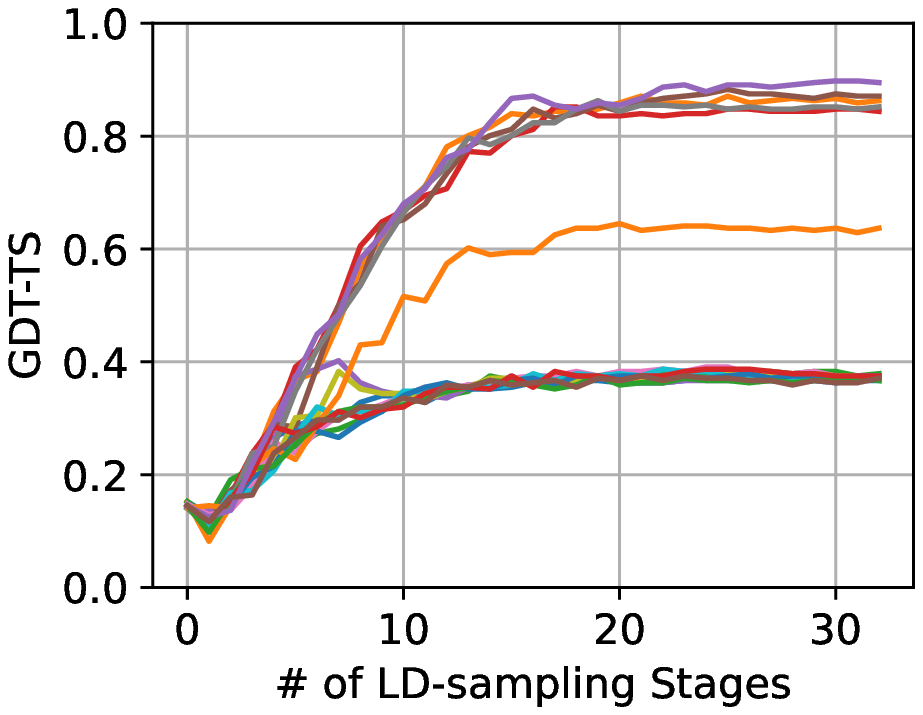}
  \caption{GDT-TS w/o HIRM}
\end{subfigure}
\begin{subfigure}{.24\textwidth}
  \includegraphics[width=\linewidth]{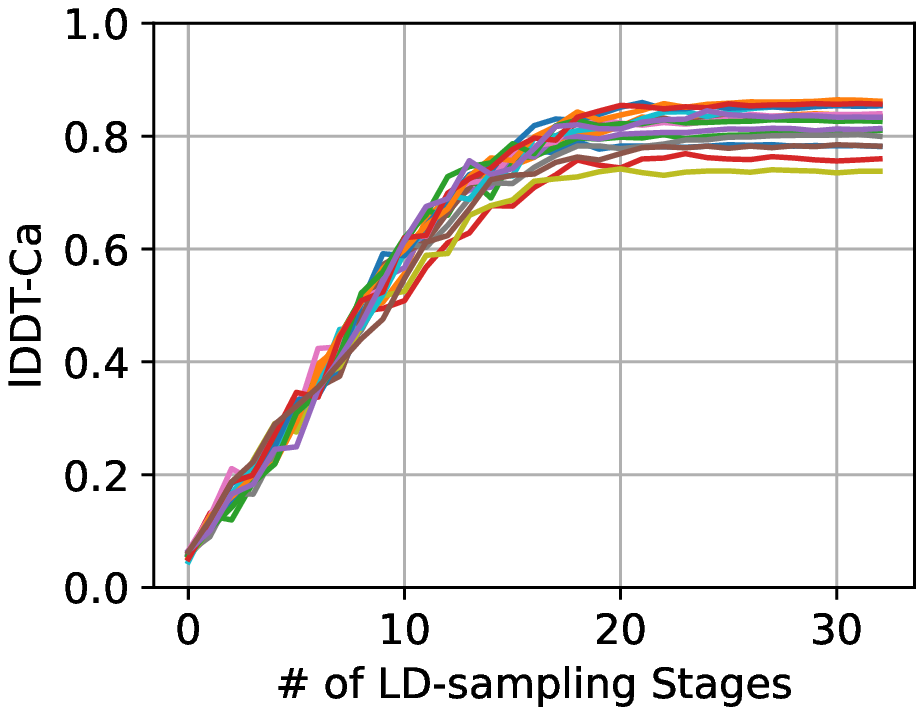}
  \caption{lDDT-Ca w/ HIRM}
\end{subfigure}
\begin{subfigure}{.24\textwidth}
  \includegraphics[width=\linewidth]{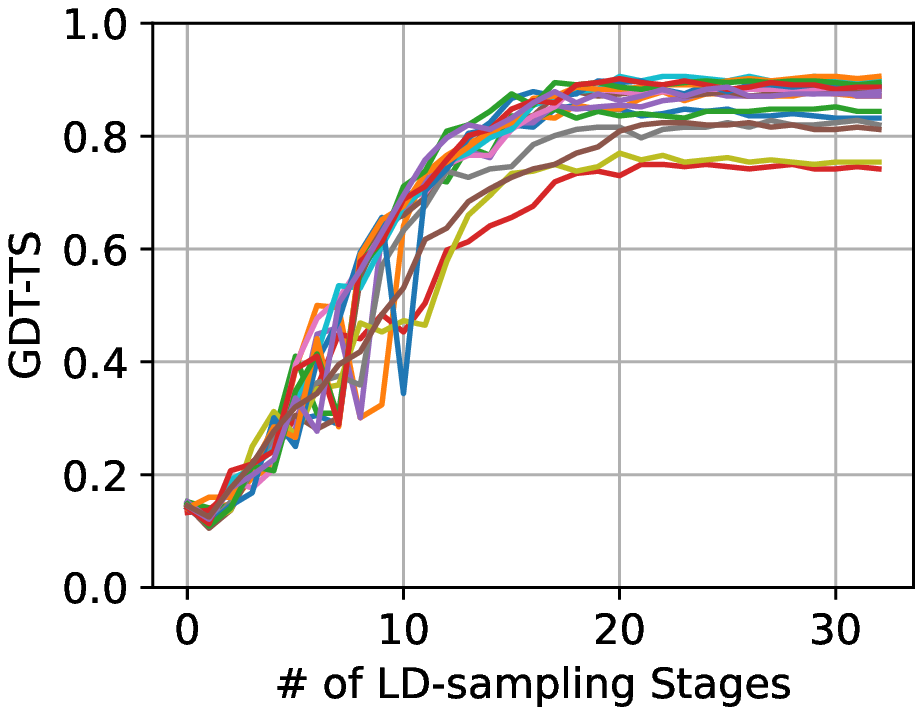}
  \caption{GDT-TS w/ HIRM}
\end{subfigure}
\hspace*{\fill} \\
\hspace*{\fill}
\begin{subfigure}{.24\textwidth}
  \includegraphics[width=\linewidth]{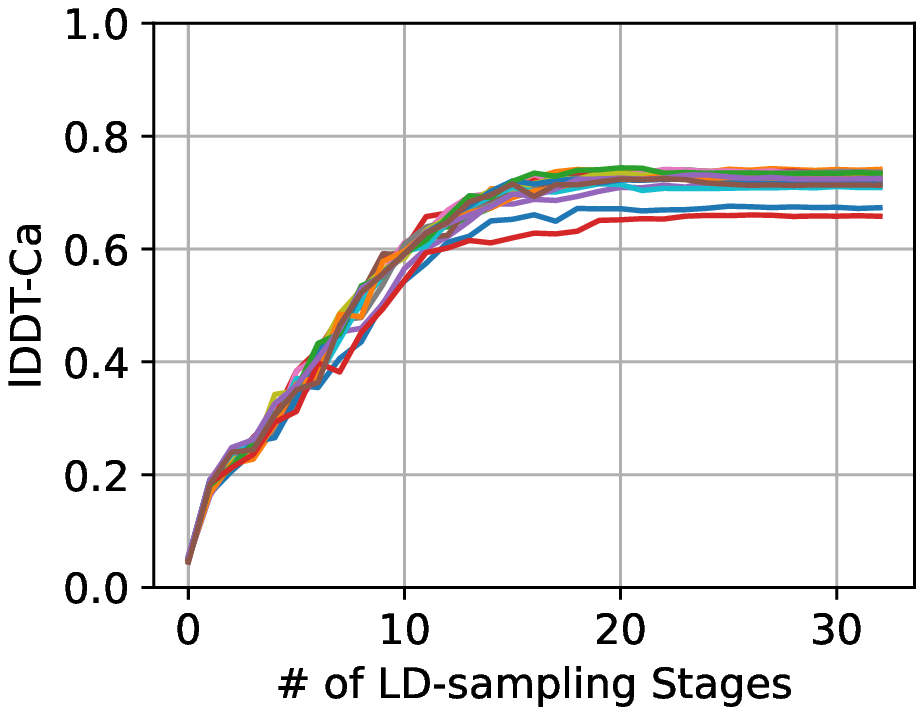}
  \caption{lDDT-Ca w/o HIRM}
\end{subfigure}
\begin{subfigure}{.24\textwidth}
  \includegraphics[width=\linewidth]{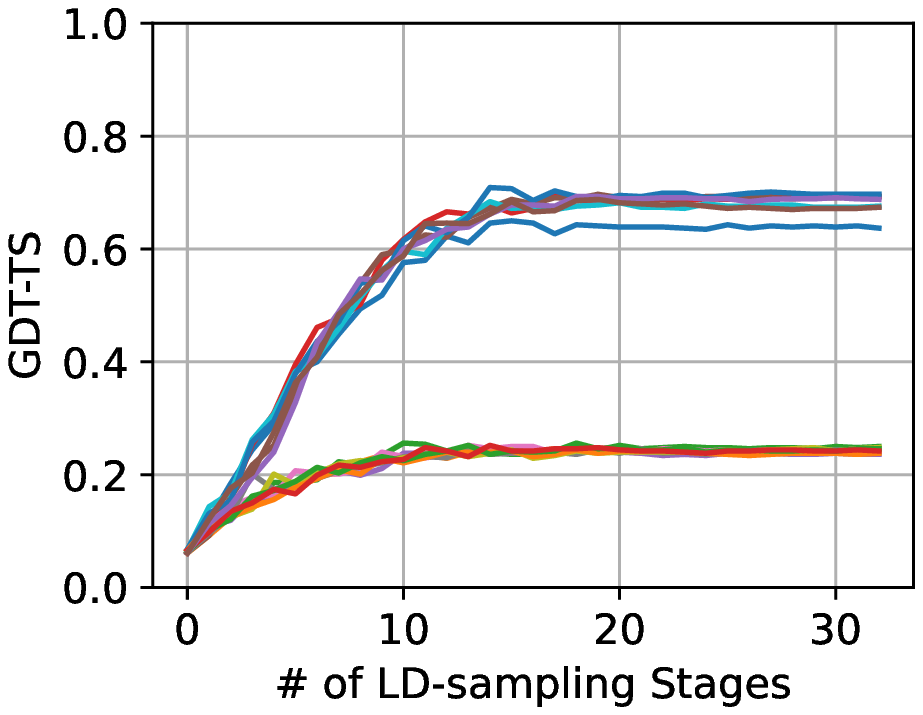}
  \caption{GDT-TS w/o HIRM}
\end{subfigure}
\begin{subfigure}{.24\textwidth}
  \includegraphics[width=\linewidth]{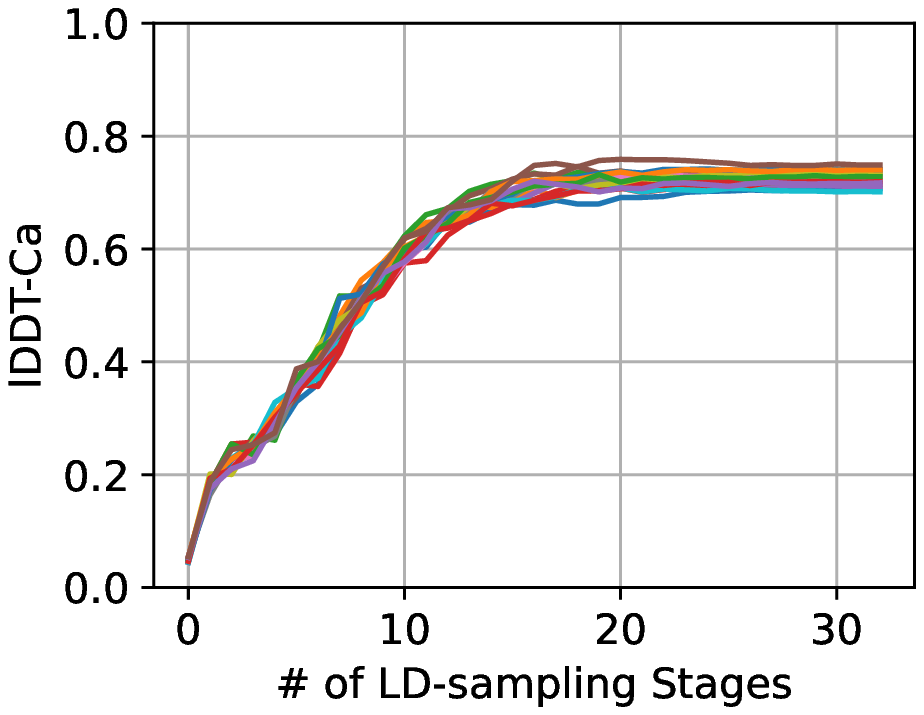}
  \caption{lDDT-Ca w/ HIRM}
\end{subfigure}
\begin{subfigure}{.24\textwidth}
  \includegraphics[width=\linewidth]{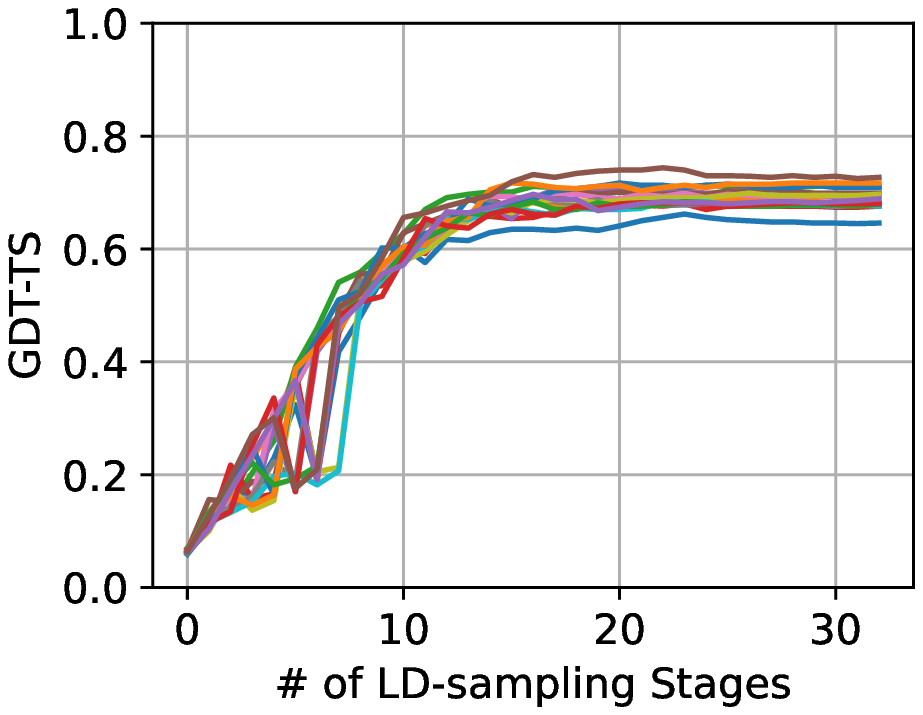}
  \caption{GDT-TS w/ HIRM}
\end{subfigure}
\hspace*{\fill} \\
\hspace*{\fill}
\begin{subfigure}{.24\textwidth}
  \includegraphics[width=\linewidth]{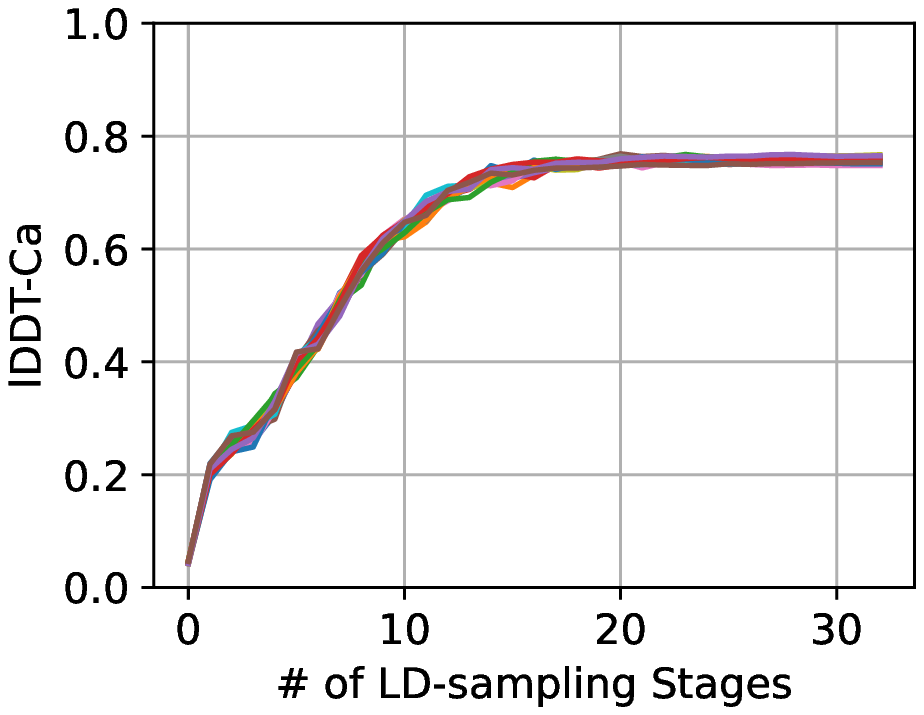}
  \caption{lDDT-Ca w/o HIRM}
\end{subfigure}
\begin{subfigure}{.24\textwidth}
  \includegraphics[width=\linewidth]{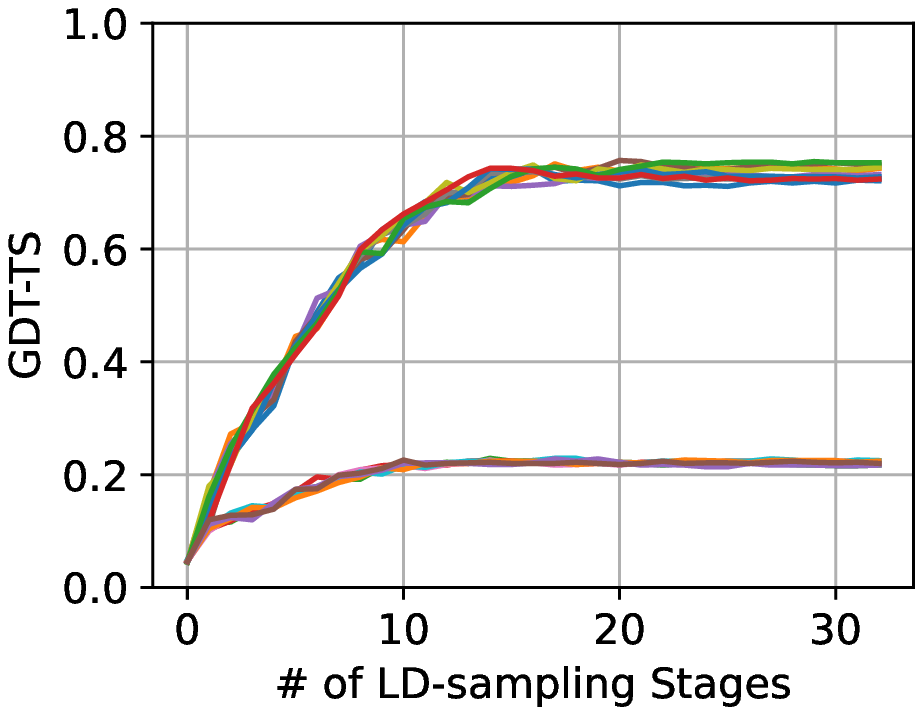}
  \caption{GDT-TS w/o HIRM}
\end{subfigure}
\begin{subfigure}{.24\textwidth}
  \includegraphics[width=\linewidth]{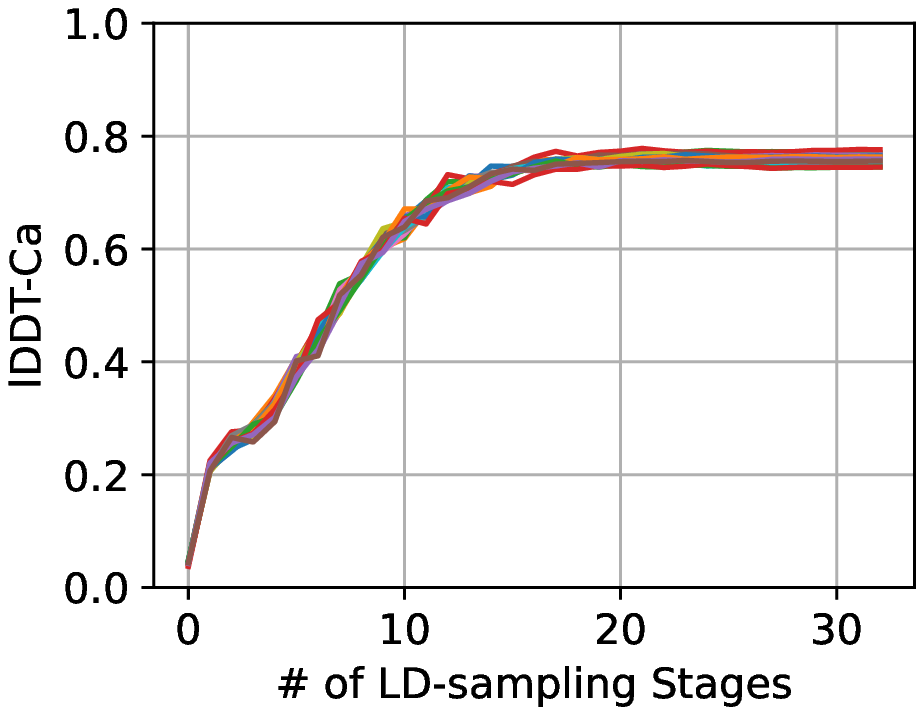}
  \caption{lDDT-Ca w/ HIRM}
\end{subfigure}
\begin{subfigure}{.24\textwidth}
  \includegraphics[width=\linewidth]{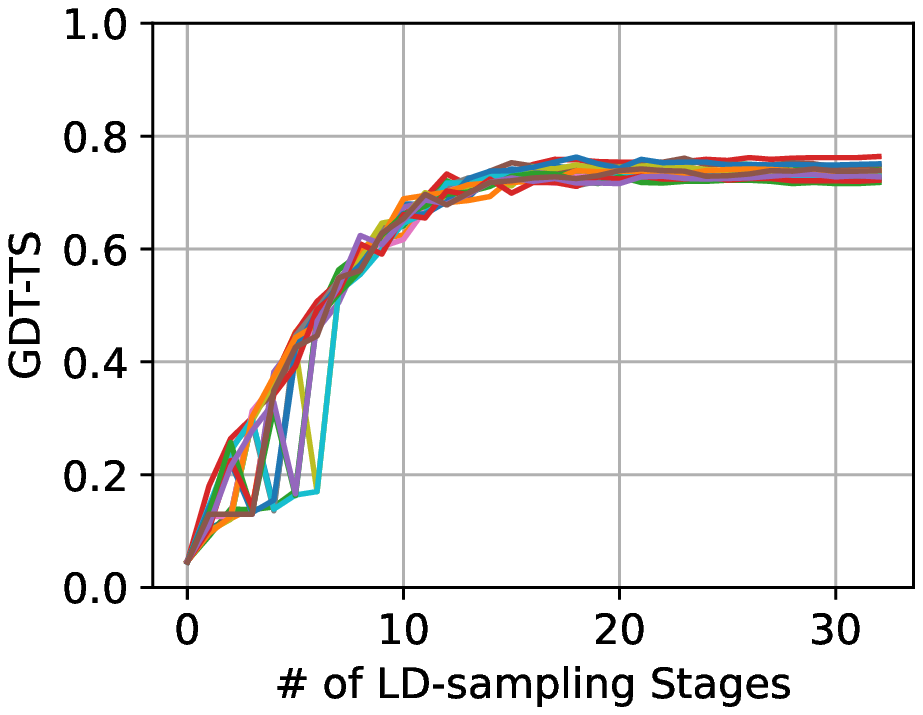}
  \caption{GDT-TS w/ HIRM}
\end{subfigure}
\hspace*{\fill}
\caption{Comparison on lDDT-Ca and GDT-TS scores during the structure optimization process, with or without the handedness issue resolving module (HIRM) enabled. Top: 1AHO-A00; medium: 2BSJ-A00; bottom: 3A2Z-A00.}
\label{fig:lddt_gdt_handedness}
\end{figure}

As mentioned earlier, since one valid distance matrix corresponds to two possible structures (mirror images of each other), \ours{} explicitly resolves the handedness issue based on the dihedral angle distribution defined on $C_{\alpha}$ atoms. Here, we verify whether this scheme can indeed prevent \ours{} from producing structure predictions with incorrect handedness. Because the lDDT-Ca metric is invariant to mirrored structures, we adopt GDT-TS \cite{zemla2003lga} as an additional evaluation metric for \ours{}'s optimized structures. In Figure \ref{fig:lddt_gdt_handedness}, we report lDDT-Ca and GDT-TS scores during the structure optimization process, with or without the handedness issue resolving module enabled.

From Figure \ref{fig:lddt_gdt_handedness}, we discover that although the usage of handedness issue resolving module does not affect how lDDT-Ca scores change throughout the optimization process, it has a significant impact on GDT-TS scores, which is handedness sensitive. When the handedness issue resolving module is disabled, around half of predicted structures' GDT-TS scores are notably lower, indicating an incorrect handedness in these structures. This phenomenon is not observed when the handedness issue resolving module is employed, suggesting that this module can indeed fix incorrect handedness in predicted structures.

\section{Conclusions and Future Work}

In this paper, we present \ours{}, a fully-differentiable approach for protein structure prediction. This offers an alternative choice for protein structure optimization, which is previously dominated by traditional tools like Rosetta and I-TASSER. The \ours{} approach shows promising accuracy when comparing against the state-of-the-art structure optimization protocol, trRosetta. This points out a possible way to building an end-to-end framework for protein structure prediction, similar as AlphaFold2.

There are still much to explore, based on the current methodology and experimental results. SE(3)-equivariant models could be an alternative choice for the score network's architecture. The score network could be more sufficiently trained with a large-scale training dataset. Extending the structure representation from $C_{\alpha}$ atoms only to all the atoms is also worth further investigation. All of these may bring further boost to the protein structure optimization's accuracy, and make progress towards full reimplementation of AlphaFold2.

\bibliographystyle{plain}
\bibliography{reference}

\end{document}